\pdfoutput=1
\documentclass[11pt,table]{article}

\usepackage[table]{xcolor}
\usepackage{EMNLP2022}

\usepackage{times}
\usepackage{latexsym}
\usepackage{amsmath,amsfonts}
\usepackage{booktabs}
\usepackage{graphicx}
\usepackage{caption}
\usepackage{subcaption}
\usepackage[ruled,vlined]{algorithm2e}
\usepackage{algorithmicx}
\usepackage{algpseudocode}
\usepackage{makecell}
\usepackage{soul}

\usepackage{tikz}
\usepackage{diagbox}
\usepackage{multirow}
\usepackage{float}
\usepackage{tablefootnote}

\usepackage[T1]{fontenc}
\usepackage[utf8]{inputenc}

\usepackage{microtype}

\newcommand{\comment}[1]{}

\title{Factorizing Content and Budget Decisions in \\ Abstractive Summarization of Long Documents}

\author{Marcio Fonseca $\qquad$ Yftah Ziser $\qquad$ Shay B. Cohen \\
Institute for Language, Cognition and Computation \\
School of Informatics, University of Edinburgh \\
10 Crichton Street, Edinburgh, EH8 9AB \\
\medskip
\texttt{m.fonseca@ed.ac.uk},
\texttt{yftah.ziser@ed.ac.uk},
\texttt{scohen@inf.ed.ac.uk}}
  
\newcommand{\modelname}{FactorSum}

\newenvironment{itemizesquish}[2]{\begin{list}{\labelitemi}{\setlength{\itemsep}{#1}\setlength{\labelwidth}{#2}\setlength{\leftmargin}{\labelwidth}\addtolength{\leftmargin}{\labelsep}}}{\end{list}}

\begin{document}
\maketitle
\begin{abstract}
We argue that disentangling content selection from the budget used to cover salient content improves the performance and applicability of abstractive summarizers. Our method, \textsc{\modelname}\footnote{Code is available at \url{https://github.com/thefonseca/factorsum}.}, does this disentanglement by factorizing summarization into two steps through
an energy function: (1) generation of \emph{abstractive summary views} covering salient information in subsets of the input document (\emph{document views})
; (2) combination of these views into a final summary, following a budget and content guidance. This guidance may come from different sources, including from an \emph{advisor} model such as BART or BigBird, or in oracle mode -- from the reference. This factorization achieves significantly higher ROUGE scores on multiple benchmarks for long document summarization, namely PubMed, arXiv, and GovReport. Notably, our model is effective for domain adaptation. When trained only on PubMed, it achieves a 46.29 ROUGE-1 score on arXiv, outperforming PEGASUS trained in domain by a large margin. Our experimental results indicate that the performance gains are due to more flexible budget adaptation and processing of shorter contexts provided by partial document views.
\end{abstract}

\section{Introduction}

Casting summarization as a language transduction problem is convenient given the existence of powerful neural sequence-to-sequence models that produce high-quality textual outputs \citep{zhang2020pegasus,lewis2019bart}. However, this framework conflates multiple steps of the summarization decision-making process into a single feedforward step without taking into account the contextual factors involved \citep{jones1999automatic}.

One such decision depending on context factors is the quantity of information to be included in a summary, reflecting on the generated outputs' length. This factor is particularly relevant for long documents that cover many different aspects of interest for which different summaries may be suitable. For instance, in samples from summarization datasets such as PubMed and arXiv \citep{cohan2018discourse}, there are many abstracts including terse passages about the background or methods of the research. In contrast, others will add more details about those aspects. Often, those choices are due to the author's preferences and do not necessarily represent an ideal summary for the document.

Furthermore, current evaluation protocols based on n-gram overlap are sensitive to summary lengths \cite{sun2019compare}. Generating summaries that match ground-truth lengths increases performance (see Section \ref{sec:results-budget-guidance}). Thus, recent progress in summarization may be the effect of better length prediction and not the actual summarization desideratum: a reductive transformation of the source text that keeps the important information \citep{jones1999automatic}.

To address this issue, we propose to avoid budget information as a confounding factor as much as possible in sequence-to-sequence training. Instead, we treat budget decisions as \emph{extrinsic guidance} during summary generation, that is, an objective that is unrelated to the content of the documents. In this setting, the neural abstractive model is responsible for the generation of short passages (\emph{summary views}) capturing relevant topics of the input document (\emph{intrinsic importance} objective), while the \emph{extrinsic importance} objective will encourage the adherence of generated summaries to context factors such as budgets or aspect coverage.

Specifically, we formulate \textsc{\modelname}, a factorized energy-based model \citep{lecun2006tutorial} aiming to find a summary that maximizes the total importance given a source document, a reference dataset, and contextual factors such as budget and content guidance. A key piece of our model is the sampling of \emph{random document views} (and corresponding reference \emph{summary views}) which allows the abstractive model to focus on shorter summarization tasks with less influence of varying summary lengths. Also, this approach allows the processing of long documents without truncation, which is a recurring problem in summarization \citep{beltagy2020longformer,zaheer2020big}. The sampling procedure is detailed in Section \ref{sec:sampling_document_views}.

\begin{figure}
    \centering
    \includegraphics[trim={0.5cm 0.3cm 1cm 0cm},clip,width=0.48\textwidth]{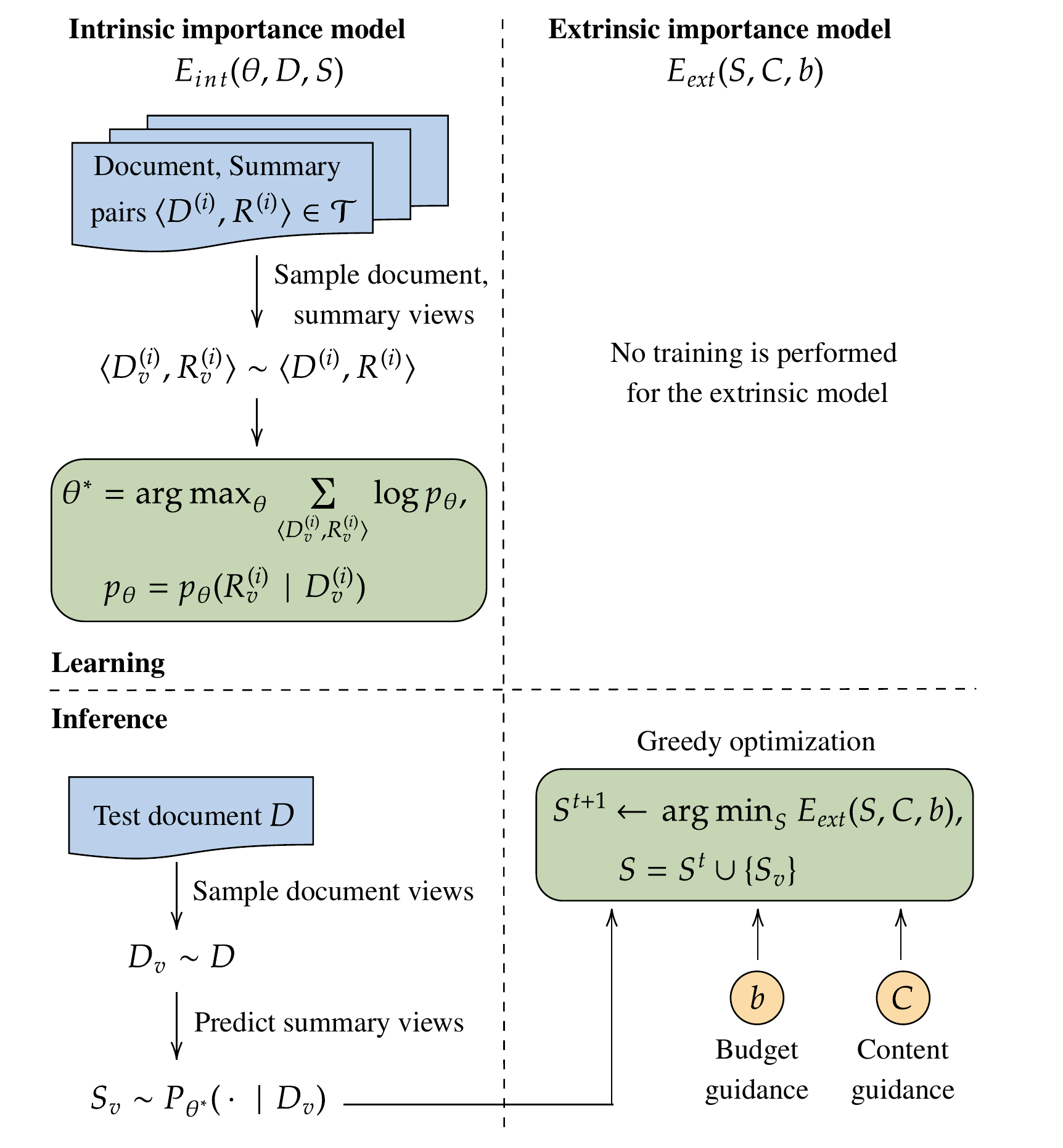}
    \caption{An overview of the summarization model. Each quadrant represents the learning/inference step of either the intrinsic or extrinsic importance model. Intrinsic learning is implemented as the usual training of a sequence-to-sequence model $p_{\theta}$, but using shorter sampled document and summary views. Optimization procedures are represented by rounded rectangles.}\label{fig:model_overview}
\end{figure}

Our model comprises two optimization procedures: learning and inference. In the learning phase, the model parameters are optimized so that summary views with important content (as informed by reference summaries) will have lower energies. In practice, this is implemented by training a neural sequence-to-sequence model to predict summary views. During inference, a greedy optimization algorithm is used to find the combination of summary views that maximize the compatibility with the target budget and other types of guidance. This process is illustrated in Figure~\ref{fig:model_overview}.

Our experimental results on the PubMed, arXiv, and GovReport summarization benchmarks \citep{cohan2018discourse, huang2021efficient} show that our approach using budget guidance alone is competitive with resource-intensive baselines such as PEGASUS \citep{zhang2020pegasus}. The results confirm that matching reference summary lengths significantly impacts ROUGE scores, often more than different modeling approaches.
We also investigate the use of existing baselines as additional guidance during summary generation. In contrast to teacher models in knowledge distillation literature \citep{hinton2015distilling}, we leverage existing model prediction during inference only, and thus, we adopt the term \emph{advisor} model to refer to our summarization guidance approach. When guided by BigBird or BART, our model obtains state-of-the-art results on PubMed, arXiv, and GovReport.

Finally, we perform domain adaptation experiments in which models trained on PubMed, arXiv, and GovReport have no access to samples from the evaluation dataset during training. Our results indicate that \textsc{\modelname} can adapt better to out-of-domain data, outperforming strong baselines trained in domain on both PubMed and arXiv. This finding suggests a good generalization capacity and is evidence that we achieved our objective to disentangle content selection from budget decisions.

\section{Intrinsic and Extrinsic Importance}
\label{sec:methods}
Since our objective is to explicitly model budget decisions, we need a definition of importance that accounts for context factors. Inspired by the information-theoretic notion of importance developed by \newcite{peyrard2018simple}, we introduce the notion of importance with respect to \emph{intrinsic} and \emph{extrinsic} semantic units.

Intrinsic semantic units are those specific to the document (e.g., salient topics) whereas extrinsic units are related to a priori external preferences or require domain knowledge and grounding that is hard to capture from the textual corpora alone. We argue that the usual setting of end-to-end supervised summarization evaluated by ROUGE \citep{lin2004rouge} optimizes for intrinsic importance. In this work, budget and content guidance provided by advisor model summaries (Section \ref{sec:extrinsic_importance}) play the role of extrinsic information.

Formally, we define the best summary $S^*$ for a document $D$ as the summary that minimizes the following factorized energy \citep{lecun2006tutorial}:
\begin{align}\label{eq:factorized_energy}
E(\theta, D, S, C, b) &= \\
& E_{int}(\theta, D, S) + E_{ext}(S,C,b), \nonumber
\end{align}
where we call $E_{int}$ and $E_{ext}$ intrinsic and extrinsic energies respectively, while $b$ denote the summary budget guidance and $C$ is a guidance content provided by an advisor model as explained in Section \ref{sec:extrinsic_importance}. 
This energy point of view allows us to unify the notion of extrinsic and intrinsic semantic units, and present the duality of the energy functions with respect to learning and inference.

Furthermore, the factorization of the total energy function makes the problem more tractable and leads to the following advantages:
\begin{itemizesquish}{-0.3em}{0.5em}
    \item Model components can be changed or replaced more cost-effectively. For example, adding more components to the extrinsic objective would not require retraining the intrinsic importance model.
    \item Issues with differentiability of the extrinsic guideline loss with respect to the summary views generator parameters are avoided.
    \item More complex inference procedures that just feedforward computation are possible.
\end{itemizesquish}

An overview of the model components and the summary inference procedure is represented in Figure \ref{fig:model_overview}. Given a document $D$, $n_d$ document views are generated, each covering a random subset of sentences from the original document (Section \ref{sec:sampling_document_views}). Then, the intrinsic importance model generates \emph{summary views} $S_v$, each partially covering salient content from the original document $D$ (Section \ref{sec:intrinsic_importance}). Finally, the extrinsic importance model will optimize the final summary $S$ so that it maximizes the alignment of the content to a target budget and content guidance. In the following sections, we detail the model components as well as the training and inference procedures.

\subsection{Sampling Document Views}
\label{sec:sampling_document_views}
Our model requires multiple summary proposals (or views) to allow the optimization of the extrinsic energy (Eq.~\ref{eq:factorized_energy}). One further motivation for using document views is that it allows the intrinsic encoder-decoder model to focus on shorter sequences, which makes the less affected by truncation issues. To generate multiple \emph{views} for the same document, we implement the following steps:
\begin{itemizesquish}{-0.3em}{0.5em}
    \item From a document $D$, we generate a random sample of sentences, which we call a \emph{document view} $D_v$. The number of sentences in $D_v$ is controlled by the sampling factor parameter $s_f \in [0,1]$, so that $\mathrm{n\_sents}(D_v) \approx s_f \cdot \mathrm{n\_sents}(D)$.
    \item Also from $D$, we extract oracle sentences $o_i$ corresponding to each sentence $r_i$ in the reference summary $R$. We choose as oracle the sentences that maximize the sum of ROUGE-1 and ROUGE-2 F1 scores: $o_i = \operatorname*{argmax}_{s \in D} \mathrm{ROUGE\text{\_}1}(s, r_i) + \mathrm{ROUGE\text{\_}2}(s, r_i)$.
    \item For each oracle sentence $o_i \in D_v$ we collect the corresponding sentence $r_i$ from the reference summary $R$. These sentences $r_i$ form the reference summary $R_v$ for the document view $D_v$. If there is no oracle sentence in the document view, the reference summary view $R_v$ is empty\footnote{Except for training, when we enforce that each document view has at least one oracle sentence.}.
\end{itemizesquish}
For each document $D$ from a training dataset $\mathcal{T}$, we repeat the sampling procedure described above $n_d$ times, yielding a new dataset $\mathcal{T'} = \{(D_v^{(i)}, R_v^{(i)})\ : i = 1, \ldots, | \mathcal{T}| \times n_d\}$ with $n_d$ times more samples than the original data. The number of samples $n_d$ and the sample fraction $s_f$ are hyperparameters that to be tuned for each dataset. For PubMed, by sampling $n_d=20$ views per document, each with 20\% of the document sentences, we obtain document views with 17.2 sentences on average, while covering 99.1\% of the original oracle sentences. In Appendix \ref{sec:document_sampling_details}, we provide statistics for different $n_d$ and $s_f$ and the heuristics we use for choosing appropriate values.

The intuition behind this sampling method is that if a sentence is relevant for the entire document, it should also be relevant in different contexts. Besides allowing the decoupled energy minimization objective, this approach also makes the input documents and corresponding summaries much shorter than the original data. Thus, our method scales to long documents without requiring specialized architectures for modeling long sequences \citep{beltagy2020longformer, zaheer2020big}. Also, in contrast to previous work, this summary sampling is domain-agnostic as it does not make any assumption about the discourse structure of the document \citep{dong-etal-2021-discourse, gidiotis2020divide}.

\subsubsection{Intrinsic Importance Model}
\label{sec:intrinsic_importance}
Powerful sequence-to-sequence models such as PEGASUS \citep{zhang2020pegasus} and BART \citep{lewis2019bart} are trained to estimate the probability of a sequence of tokens given a document by minimizing cross-entropy with respect to the data distribution. We hypothesize that these models are good candidates to fulfill the intrinsic importance objective, as described below.

\paragraph{Learning}
Given the training dataset $\mathcal{T'}$ consisting of document views $D^{(i)}_v$ and reference summary views $R^{(i)}_v$ (Section \ref{sec:sampling_document_views}), we define the intrinsic loss as a negative log-likelihood functional:
\begin{align} \label{eq:intrinsic_loss}
L(E_{int}, \mathcal{T}') &= \frac{1}{|\mathcal{T}'|} \sum_{i=1}^{|\mathcal{T}'|} \underbrace{L(R^{(i)}_v, E_{int}(\theta, D^{(i)}_v, \mathcal{S}))}_{-\displaystyle {\log p_\theta(R^{(i)}_v | D^{(i)}_v)}} \nonumber 
%    &= \frac{1}{|\mathcal{T}'|} \sum_{i=1}^{|\mathcal{T}'|} -\log ,
\end{align}
where $p_\theta(R^{(i)}_v | D^{(i)}_v)$ is a distribution over the possible summaries $\mathcal{S}$, specifically a sequence-to-sequence neural network model \citep{lewis2019bart}. During learning, we find the parameters $\theta^*$ that minimize the loss above.

\paragraph{Inference} 
The summary generation is performed as usual in sequence-to-sequence models via beam search decoding \citep{sutskever2014sequence}. We sample summary views by generating a summary conditioned to the document views:
\begin{align}
S^{(i)}_v \sim p_{\theta^*}(\ \cdot\ | D^{(i)}_v).
\end{align}
We assume these summary views are samples from low-energy regions of $E_{int} = -\log p_{\theta^*}(S^{(i)}_v | D^{(i)}_v)$, thus contributing to the minimization of the factorized energy (Eq.~\ref{eq:factorized_energy}).

\subsubsection{Extrinsic Importance Model}
\label{sec:extrinsic_importance}
The extrinsic importance energy function $E_{ext}$ measures the compatibility between the summary, the guidance budget $b$, and the guidance content $C$. Thus, the optimal summary $S^*$ is defined as:
\begin{align}
S^* &= \arg\min_{S} E_{ext}(S,C,b).
\end{align}
The extrinsic energy is defined in terms of the squared deviation with respect to the guidance budget and ROUGE-1 score between the generated summary and the guidance content:
\begin{align} \label{eq:extrinsic_energy}
E_{ext}(&S,C,b) \\ \nonumber
    &= \alpha\ (|S| / b - 1)^2 - \beta\  \mathrm{ROUGE\text{\_}1}(S, C),
\end{align}
where $|S|$ denotes the length of the summary in words, the content C is a summary provided by an advisor model. The hyperparameters $\alpha$ and $\beta$ weight the contribution of each guidance signal. In our experiments, we use $\alpha=\beta=1.0$ and, as advisor models, PEGASUS \citep{zhang2020pegasus} and BigBird \citep{zaheer2020big}.

In our implementation, there is no learning step for the extrinsic importance model. For inference, we design a greedy algorithm to minimize the energy as detailed in Algorithm \ref{alg:greedy_summary}. Let $V_D$ be the set of $n_d$ summary views for the document $D$. Starting from the initial condition $S = \emptyset$, the procedure selects the summary view $S_v \in V_D$ that minimizes the energy $E_{ext}(S \cup \{S_v\},C,b)$. The view $S_v$ is added to the summary if it satisfies the following additional conditions:
\begin{itemizesquish}{-0.3em}{0.5em}
    \item \textbf{Non-redundancy}: the view $S_v$ cannot be redundant with respect to the current summary $S$. We consider as redundant a summary view that has a (word-level) normalized Levenshtein distance\footnote{We use the \texttt{textdistance} library:\\ \url{https://github.com/life4/textdistance}} \citep{levenshtein1966binary} to any sentence in $S$ lower than a threshold $t=0.4$\footnote{The redundancy threshold was manually tuned by inspecting sample outputs from the validation set.} (\texttt{is\_redundant} function in Algorithm \ref{alg:greedy_summary}).
    \item \textbf{Energy reduction}: $S \cup \{S_v\}$ must have a lower energy than the current best summary $S^*$. After $p$ iterations without improvement, the algorithm returns the current best summary $S^*$. Unless otherwise stated, this patience parameter is set to $p=n_d$, which means the algorithm iterates over all available views.
\end{itemizesquish}
When $\beta = 0$ in Eq.~\ref{eq:extrinsic_energy} (no content guidance), each step of the greedy algorithm adds the longer summary view that satisfies the non-redundancy condition above, except for the last step, when a shorter 
view may better match the budget guidance. We provide further details on pre- and postprocessing summary views in Appendix \ref{sec:inference_details}.
\DecMargin{2em}
\begin{algorithm}[t]
    \hfill
    \begin{minipage}{0.87\linewidth}
    \small
    \SetAlgoLined
    \KwIn{$V_D, C, b, t, p$}
    \KwOut{$S^*$} 
    $S \gets \emptyset, S^* \gets \emptyset$\;
    $i \gets 0$\;
    \While{$V_D \neq \emptyset ~ \mathbf{and} ~ i \leq p$}{
        $S^*_v \gets \arg\min_{S_v \in V_D} E_{ext}(S \cup \{S_v\}, C, b)$
    
        \uIf{$E_{ext}(S \cup \{S^*_v\}, C, b) > E_{ext}(S^*, C, b)$}{
            $i \gets i + 1$\;
            $S \gets S \cup \{S^*_v\}$\;
        } \uElseIf{$\mathbf{not} ~ is\_redundant(S,S^*_v, t)$} {
            $i \gets 0$\;
            $S \gets S \cup \{S^*_v\}$\;
            $S^* \gets S$\;
        }
        $V_D \gets V_D \setminus \{S^*_v\}$\;
    }
    \end{minipage}

    \caption{Greedy summary generation. Input parameters are the set of summary views $V_D$ for document $D$, content guidance $C$, budget guidance $b$, redundancy threshold $t$, and patience $p$. See the "Non-redundancy" paragraph in Section \ref{sec:extrinsic_importance} for a discussion about the \texttt{is\_redundant} function.}\label{alg:greedy_summary}
\end{algorithm}

\section{Experimental Setup}
\label{sec:experiments}
We turn to discuss the experimental setup to assess the effectiveness of the factorized model proposed in Section \ref{sec:methods}. We specify details about datasets, baselines, and the evaluation protocol. (See Appendices \ref{sec:training_details} and \ref{sec:inference_details} for further implementation details).

\paragraph{Datasets}
Our experiments are performed on the PubMed and arXiv datasets, consisting of documents extracted from the homonymous scientific repositories \citep{cohan2018discourse}. To further test the generalization capacity of the model, we also perform the experiments on GovReport, a dataset containing long reports published by
U.S. Government Accountability Office (GAO; \citealt{huang2021efficient}). The only preprocessing applied is to filter out documents with empty articles or summaries, which lead to the training, validation, and test splits shown in Table \ref{tab:dataset_statistics}. We do not truncate the articles or their abstracts.
\begin{table}
  \setlength\tabcolsep{2.5pt}
  \centering
  \begin{tabular}{lccccc}
    \toprule
    \multicolumn{1}{l}{\multirow{2}{1cm}{\centering \textbf{Dataset}}} & \multicolumn{3}{c}{ \textbf{Samples}} & \multicolumn{2}{c}{ \textbf{Summaries}} \\
    \cmidrule(r){2-4} \cmidrule(r){5-6}
     &  \textbf{Train} &  \textbf{Val} &  \textbf{Test} &  \textbf{Sents} &  \textbf{Words} \\
    \midrule
    PubMed & 119,920 & 6,631 & 6,658 & 6.8 & 204.8 \\
    arXiv & 202,917 & 6,436 & 6,440 & 12.6 & 292.6 \\
    GovReport & 17,517 & 973 & 973 & 17.6 & 546.0  \\
    \bottomrule
  \end{tabular}
  \caption{Key statistics for the summarization datasets. "Sents" and "Words" denote the average number of words and sentences in the summaries (training split).}\label{tab:dataset_statistics}
\end{table}

\paragraph{Evaluation}
We evaluate our models using the ROUGE F-measure metric \citep{lin2004rouge}, with the implementation used by \newcite{zhang2020pegasus}\footnote{\url{https://github.com/google-research/pegasus}}.

\paragraph{Baseline models}
We use the following summarization baselines, for which implementations and pre-trained models are publicly available:
\begin{itemizesquish}{-0.3em}{0.5em}
    \item \textbf{PEGASUS} \citep{zhang2020pegasus}, an encoder-decoder transformer-based model that uses a specialized pre-training task of predicting entire masked sentences and achieves strong performance across several datasets.
    \item \textbf{BigBird} \citep{zaheer2020big}, a model based on a sparse attention mechanism that allows transformer-based models to process up to 8 times longer sequences efficiently.
    \item \textbf{BART} \citep{lewis2019bart}, a transformer-based denoising autoencoder that has strong performance on text generation tasks. We train our own version of BART-large on GovReport with a longer maximum target length of 768 tokens.
\end{itemizesquish}
Also, we add results for the following abstractive systems: DANCER \citep{gidiotis2020divide}, HEPOS \citep{huang2021efficient}, DYLE \citep{mao_dyle_2022}, and \textsc{Summ$^N$} \citep{zhang_summn_2022}.

\section{Results}
\label{sec:results}
In this section, we analyze the contribution of different guidance factors on summarization performance by controlling summary budget and content guidance. We continue by conducting an ablation study, examining the summary views and the greedy summary generation contributions to the overall performance. Finally, we discuss domain adaptation results. Sample summaries are provided in Appendix \ref{sec:sample_summaries}.

% =======================

\begin{table*}[h]
  \centering
  \setlength\tabcolsep{3.5pt}
  \begin{tabular}{lc|cccc|cccc|cccc}
    \toprule
    \multicolumn{2}{c|}{\multirow{2}{2.2cm}{\centering \textbf{Model}}} & \multicolumn{4}{c|}{\textbf{PubMed}} & \multicolumn{4}{c|}{\textbf{arXiv}} & \multicolumn{4}{c}{\textbf{GovReport}} \\
    \cmidrule(r){3-6} \cmidrule(r){7-10} \cmidrule(r){11-14} 
    &  & \textbf{R-1} & \textbf{R-2} & \textbf{R-L} & \textbf{Len} & \textbf{R-1} & \textbf{R-2} & \textbf{R-L} & \textbf{Len} & \textbf{R-1} & \textbf{R-2} & \textbf{R-L} & \textbf{Len} \\
    \toprule
    \multicolumn{14}{c}{\textbf{Previous work}} \\
    \midrule
    \multicolumn{2}{l|}{DANCER$\dagger$} & 46.34 & 19.97 & 42.42 & - & 45.01 & 17.60 & 40.56 & - & - & - & - & - \\
    \multicolumn{2}{l|}{HEPOS$\dagger$} & \textbf{48.12} & \textbf{21.06} & 42.72 & - & 48.24 & 20.26 & 41.78 & - & 56.86 & 22.62 & 53.82 & - \\
    \multicolumn{2}{l|}{DYLE$\dagger$} & - & - & - & - & 46.41 & 17.95 & 41.54 & - & \textbf{61.01} & \textbf{28.83} & \textbf{57.82} & - \\
    \multicolumn{2}{l|}{\textsc{Summ$^N$}$\dagger$} & - & - & - & - & - & - & - & - & 56.77 & 23.25 & 53.90 & - \\
    \multicolumn{2}{l|}{PEGASUS} & 43.83 & 18.72 & 40.29 & 180 & 43.06 & 16.39 & 38.65 & 168 & - & - & - & - \\
    \multicolumn{2}{l|}{BigBird} & 45.48 & 19.92 & 41.81 & 185 & 46.15 & 18.60 & 41.46 & 164 & - & - & - & - \\
    \multicolumn{2}{l|}{BART-large} & - & - & - & - & - & - & - & - & 52.82 & 19.12 & 49.99 & 596 \\
    \midrule
    \multicolumn{2}{c|}{\textbf{Guidance}} & \multicolumn{12}{c}{} \\
    \cmidrule(r){1-2}
    \textbf{Budget} & \textbf{Content} & \multicolumn{9}{c}{\textbf{\textsc{\modelname} - no content guidance}} \\
    \midrule
    Oracle & - & 47.37 & 19.10 & 43.27 & 208 & 48.87 & 18.83 & 43.96 & 167 & 59.80 & 24.13 & 56.12 & 651 \\
    Fixed & - & 45.41 & 18.66 & 41.63 & 206 & 47.22 & 18.60 & 42.61 & 165 & 58.77 & 23.99 & 55.19 & 650 \\
    Model & - & 44.64 & 17.98 & 40.76 & 185 & 46.40 & 18.21 & 41.85 & 164 & 57.18 & 23.34 & 53.66 & 638 \\
    \midrule
    \multicolumn{14}{c}{\textbf{\textsc{\modelname} - content guidance}} \\
    \midrule
    Oracle & Lead & 48.31 & 19.99 & 44.35 & 208 & 49.69 & 19.32 & 44.85 & 166 & 60.73 & 25.24 & 57.20 & 650 \\
    Fixed & Lead & 46.27 & 19.29 & 42.57 & 205 & 48.05 & 19.05 & 43.49 & 165 & 59.67 & 25.02 & 56.22 & 649 \\
    Fixed & Model & 47.50 & 20.33 & \underline{\textbf{43.76}} & 205 & \underline{\textbf{49.32}} & \underline{\textbf{20.27}} & \underline{\textbf{44.76}} & 165 & 60.10 & 25.28 & 56.65 & 648 \\
    Model & Model & 47.34 & 20.31 & \underline{43.52} & 185 & 48.74 & \underline{20.12} & 44.19 & 164 & 58.78 & 24.87 & 55.37 & 638 \\
    \bottomrule
  \end{tabular}
  \caption{ROUGE F1 scores and average words per summary on the test sets for different types of guidance during inference. \emph{Lead} guidance is the first $k$ sentences from the source document (Section \ref{sec:results_content_guidance}). Model guidance is provided by BART-large for GovReport and BigBird for PubMed and arXiv. The choice of budget guidance values is described in Appendix \ref{sec:inference_details} and validation scores are provided in Appendix  \ref{sec:validation_results}. Results for models marked with $\dagger$ are taken from the original publications. \underline{Underlined results} are statistically  equivalent to the best methods ($p<0.05$).}\label{tab:experiment_results}
\end{table*}

\begin{table*}
  \centering
  \setlength\tabcolsep{3.8pt}
  \begin{tabular}{l|cccc|cccc|cccc}
    \toprule
    \multirow{2}{3em}{\diagbox[width=7.5em,height=3em]{\textbf{Training}}{\textbf{Evaluation}}} & \multicolumn{4}{c|}{\textbf{PubMed}} & \multicolumn{4}{c|}{\textbf{arXiv}} & \multicolumn{4}{c}{\textbf{GovReport}} \\
    \cmidrule(r){2-5} \cmidrule(r){6-9} \cmidrule(r){10-13}
    & \textbf{R-1} & \textbf{R-2} & \textbf{R-L} & \textbf{Len} & \textbf{R-1} & \textbf{R-2} & \textbf{R-L} & \textbf{Len} & \textbf{R-1} & \textbf{R-2} & \textbf{R-L} & \textbf{Len} \\
    \toprule
    \multicolumn{13}{c}{\textbf{End-to-end baseline} (BigBird for PubMed and arXiv; BART-large for GovReport)} \\
    \midrule
    PubMed & \cellcolor{blue!10}45.48 & \cellcolor{blue!10}19.92 & \cellcolor{blue!10}41.81 & \cellcolor{blue!10}185 & 42.33 & 15.16 & 38.09 & 161 & 19.35 & 3.57 & 18.10 & 222 \\
    arXiv & 39.47 & 14.95 & 35.77 & 177 & \cellcolor{blue!10}46.15 & \cellcolor{blue!10}18.60 & \cellcolor{blue!10}41.46 & \cellcolor{blue!10}164 & 16.61 & 2.25 & 15.09 & 352 \\
    GovReport\hspace{2.33em} & 37.18 & 11.10 & 33.96 & 203 & 35.11 & 8.94 & 31.67 & 203 & \cellcolor{blue!10}52.82 & \cellcolor{blue!10}19.12 & \cellcolor{blue!10}49.99 & \cellcolor{blue!10}596 \\
    \midrule
    \multicolumn{13}{c}{\textbf{\textsc{\modelname} - fixed budget, no content guidance}} \\
    \midrule
    PubMed & \cellcolor{blue!10}45.41 & \cellcolor{blue!10}18.66 & \cellcolor{blue!10}41.63 & \cellcolor{blue!10}206 & 44.61 & 15.88 & 40.16 & 165 & 42.49 & 15.07 & 39.92 & 350 \\
    arXiv & 44.40 & 16.87 & 40.51 & 209 & \cellcolor{blue!10}47.22 & \cellcolor{blue!10}18.60 & \cellcolor{blue!10}42.61 & \cellcolor{blue!10}165 & \underline{\textbf{48.75}} & \underline{\textbf{18.07}} & \underline{\textbf{45.94}} & 414 \\
    GovReport & 39.67 & 12.63 & 35.37 & 213 & 38.34 & 10.74 & 33.72 & 167 & \cellcolor{blue!10}58.77 & \cellcolor{blue!10}23.99 & \cellcolor{blue!10}55.19 & \cellcolor{blue!10}650 \\
    \midrule
    \multicolumn{13}{c}{\textbf{\textsc{\modelname} - fixed budget and content guidance}} \\
    \multicolumn{13}{c}{(BigBird guidance for PubMed and arXiv; BART-large guidance for GovReport)} \\
    \midrule
    PubMed & \cellcolor{blue!10}47.50 & \cellcolor{blue!10}20.33 & \cellcolor{blue!10}43.76 & \cellcolor{blue!10}205 & \underline{\textbf{46.29}} & \underline{\textbf{17.13}} & \underline{\textbf{41.86}} & 166 & 42.24 & 15.03 & 39.68 & 344 \\
    arXiv & \underline{\textbf{45.87}} & \underline{\textbf{18.10}} & \underline{\textbf{42.02}} & 210 & \cellcolor{blue!10}49.32 & \cellcolor{blue!10}20.27 & \cellcolor{blue!10}44.76 & \cellcolor{blue!10}165 & \underline{48.65} & \underline{18.03} & \underline{45.85} & 410 \\
    GovReport & 41.27 & 14.01 & 37.10 & 211 & 40.00 & 11.85 & 35.47 & 176 & \cellcolor{blue!10}60.10 & \cellcolor{blue!10}25.28 & \cellcolor{blue!10}56.65 & \cellcolor{blue!10}648 \\
    \bottomrule
  \end{tabular}
  \caption{ROUGE F1 scores and average words per summary for the domain adaptation experiments. Models trained on PubMed, arXiv, and GovReport samples (rows) are used to summarize articles from the other dataset test splits (columns). The choice of budget guidance values is described in Appendix \ref{sec:inference_details}. \colorbox{blue!10}{Shaded scores} are in-domain results from Table \ref{tab:experiment_results}. \underline{Underlined results} are statistically equivalent to the best cross-domain scores ($p<0.05$).}\label{tab:domain_adaptation}
\end{table*}

\subsection{Effects of Budget Guidance}
\label{sec:results-budget-guidance}
To test the impact of budgets on the summarization performance we provide three types of guidance: 
\begin{itemizesquish}{-0.3em}{0.5em}
    \item \textbf{Fixed}: the budget guidance is set at a fixed value for all summaries. The budget is 205, 165, and 648 words, which are the average summary lengths in the validation sets of PubMed, arXiv, and GovReport respectively.
    \item \textbf{Oracle}: the model uses the reference summary length as the budget guidance.
    \item \textbf{Model-based}: the model uses the length of the summary produced by an advisor model (BART for GovReport and BigBird for PubMed/arXiv) as budget guidance.
\end{itemizesquish}
To fairly compare the different models, we use an additive budget correction so that the average number of tokens in system summaries is close to the average length of reference summaries from the validation set (Table \ref{tab:budget_guidance}). The average summary lengths for each model is presented in Tables \ref{tab:experiment_results} and \ref{tab:domain_adaptation}. Also, we provide ROUGE scores for varying budget guidance values in Appendix \ref{sec:budgets_fmeasure}.

\paragraph{Fixed budget}
The simpler version of \textsc{\modelname} is only guided by a fixed budget and its performance is competitive with most baselines, including PEGASUS and BigBird (Table \ref{tab:experiment_results}, "no content guidance"). It is important to note that the intrinsic importance model is based on a \texttt{bart-base} model with 139M parameters, which is 4 times smaller than PEGASUS and BigBird.

\paragraph{Oracle budgets} 
The second section of Table \ref{tab:experiment_results} shows that for \textsc{\modelname} with no content guidance, having access to the oracle lengths improves the scores by about 2, 1.6, and 1 ROUGE-1 on PubMed, arXiv, and GovReport respectively. We observe a similar effect with content guidance (third section of Table \ref{tab:experiment_results}). These results agree with our hypothesis that the impact of budgets on ROUGE is significant and often larger than the differences between different modeling approaches.

\paragraph{Model-based budgets} 
One may argue that inferring summary lengths is part of the task. Thus, we also test how summary lengths provided by BART and BigBird affect the summarization performance. For all datasets, we observe that using model budget guidance is detrimental to the scores compared to fixed budget guidance (second and third section of Table~\ref{tab:experiment_results}). These results suggest that summary lengths are hard to predict from the source documents, and highlights the potential benefits of divorcing content selection and budget optimization.

\subsection{Effects of Content Guidance}
\label{sec:results_content_guidance}
We also examine the impact of different types of content guidance. The first sort of guidance, \emph{Lead}, takes the first $k$ sentences from the source article. We choose the lowest $k$ so that the guidance text has at least the number of words as the fixed target budget for each dataset (see Section \ref{sec:results-budget-guidance}). This content guidance improves the scores by $\sim$0.8 (PubMed and arXiv) and $\sim$0.9 (GovReport) ROUGE-1 points over the model without guidance. Notably, \textsc{\modelname} with \emph{Lead} guidance achieves 48.05 ROUGE-1 on arXiv and 59.67 ROUGE-1 on GovReport \emph{without relying on predictions from strong baselines}. We can further improve performance by providing content guidance from BigBird and BART, leading to strong performance on all datasets (Table \ref{tab:experiment_results}). From these empirical results, we conclude that content guidance is a simple and effective method to turn strong sequence-to-sequence baselines into more flexible summarization systems. It should be possible to add more types of guidance to adapt the summaries to specific needs such as topic coverage.

\subsection{Ablation Study}

Our method comprises two main components, a document and summary views and a ranker extracting the most salient information from those summaries using budget guidance. To shed some light on the contribution of each component to the overall performance, we conduct two ablation analyses. First, to better understand the inherent potential of the summary views, we feed them to the \textsc{\modelname} ranker with reference (oracle) content guidance. Having reference as content guidance serves as an upper bound  for what we can achieve using the summary views. We do the same to an ensemble of PEGASUS and BigBird, which is a concatenation of their summaries, for comparison.

Second, to understand the importance of the \textsc{\modelname} ranker (see Algorithm \ref{alg:greedy_summary}), we replace it with TextRank \cite{mihalcea2004textrank}, a prominent algorithm for extractive summarization. Having two input variants (summary views and PEGASUS and BigBird ensemble) and three ranker variants (TextRank, \textsc{\modelname} - no content guidance, and \textsc{\modelname} - reference content guidance) results in six models. We use fixed-length guidance for all of the models.

Table \ref{tab:ablation_study} shows our results on the PubMed and arXiv datasets. We observe significantly higher results for \textsc{\modelname} - reference content guidance) when applied on summary views, compared to PEGASUS and BigBird Ensemble Summaries. This shows that when both inputs reach their full potential under ideal guidance, summary views are superior to an ensemble of the two strong baselines, thus containing more salient information. We observe a similar pattern when using \textsc{\modelname} - no content guidance as a ranker, showing that summary views serve as a better input to a more realistic ranker. In addition, we notice that for both types of input,  \textsc{\modelname} - no content guidance significantly outperforms TextRank, thus contributing to \textsc{\modelname} overall performance. 

\begin{table}
  \centering
  \setlength\tabcolsep{1.5pt}
  \begin{tabular}{l|ccc|ccc}
    \toprule
    \multicolumn{1}{c|}{\multirow{2}{1.2cm}{\centering \textbf{Ranker}}} &  \multicolumn{3}{c|}{\textbf{PubMed}} & \multicolumn{3}{c}{\textbf{arXiv}} \\
    \cmidrule(r){2-4} \cmidrule(r){5-7} & \textbf{R-1} & \textbf{R-2} & \textbf{R-L} & \textbf{R-1} & \textbf{R-2} & \textbf{R-L} \\
    \toprule
    \multicolumn{7}{c}{\textbf{PEGASUS + BigBird Ensemble Summaries}} \\
    \midrule
    TextRank & 43.93 & 18.33 & 38.40 & 44.15 & 16.89 & 37.32 \\
    \midrule
    FS & \underline{45.38} & \underline{\textbf{19.43}} & \underline{41.49} & 45.30 & 17.60 & 40.38 \\
    FS-Oracle & 48.90 & 21.81 & 44.76 & 49.34 & 20.13 & 43.99 \\
    \midrule
    \multicolumn{7}{c}{\textbf{Summary Views}} \\
    \midrule
    TextRank & 42.10 & 16.71 & 37.54 & 42.66 & 16.41 & 37.70 \\
    \midrule
    FS & \underline{\textbf{45.41}} & 18.66 & \underline{\textbf{41.63}} & \underline{\textbf{47.22}} & \underline{\textbf{18.60}} & \underline{\textbf{42.61}} \\
    FS-Oracle & 51.75 & 23.31 & 47.53 & 53.51 & 22.94 & 48.29 \\
    \bottomrule
  \end{tabular}
  \caption{ROUGE F1 scores on the test sets for the ensemble experiments. We compare summary predictions given by the concatenation of PEGASUS and BigBird summaries against summaries derived from \textsc{\modelname} summary views. We use two sentence rankers: an unsupervised TextRank baseline and \textsc{\modelname} extrinsic importance ranker. FS and FS-oracle use \textsc{\modelname} without content guidance and with reference summary guidance, respectively. All models use \emph{fixed budget} as described in Section \ref{sec:results-budget-guidance}. Best non-oracle results are \textbf{bold-faced}. \underline{Underlined results} are statistically equivalent to the best scores ($p<0.05$).}\label{tab:ablation_study}
\end{table}

\subsection{Domain Adaptation}
Our last experiment, unusual in the summarization literature, aims to test if a model trained on PubMed/arXiv/GovReport performs well when applied to out-of-domain (OOD) samples. Our intuition is that \textsc{\modelname} should adapt well to OOD budget distributions, whereas the intrinsic model captures domain-specific patterns with less influence from length and content position noise.

The adaptation performance for similar domains (PubMed and arXiv) is much higher than summarizing GovReport documents when trained on scientific articles. \textsc{\modelname} outperforms end-to-end baselines in all cases, especially when there is a large gap in average summary lengths between the domains. However, it can still achieve significant improvements on arXiv, for which all models output summaries with similar average lengths.

Our most important finding is that the ROUGE scores are not as severely affected as expected in this cross-domain setting. Notably, \textsc{\modelname} trained on PubMed without content guidance achieves 44.61 ROUGE-1 on arXiv, outperforming PEGASUS trained in-domain. When guided by BigBird summaries (also OOD), \textsc{\modelname} scores 46.29 ROUGE-1 on arXiv, also outperforming BigBird trained in-domain. Similar results are observed for models trained on arXiv and evaluated on PubMed.
On GovReport, \textsc{\modelname} can produce much longer summaries that cover more relevant content, which explains the substantial improvements in ROUGE scores over the end-to-end baselines. However, summaries generated by \textsc{\modelname} trained on arXiv/PubMed cannot match the average length of 650 words produced by the in-domain version. The reason for this gap is that OOD models generate summary views with a lower variety of content, which are eliminated by the redundancy control described in Section \ref{sec:extrinsic_importance}.

\section{Related Work}
\paragraph{Ranking and aggregating candidate summaries} Recent work explores the idea of sampling summary candidates that are scored to form a final summary. The SimCLS \citep{liu2021simcls} model generates several summary candidates using diverse beam search \citep{vijayakumar2016diverse}, which are ranked according to a learned evaluation function. Also, the Perturb-and-Select summarizer \citep{oved2021pass} uses similar ideas in a multi-document opinion summarization task. Instead of a diverse beam search, it performs random perturbations in the model inputs to generate candidates ranked according to a coherence model. \newcite{iso2021convex} presented COOP, a framework that improves the aggregation methods for multi-document opinion summarization representations by maximizing their input-output word overlap. Our approach differs from these models as our sampled summary views are not full summary candidates but partial views with important semantic content from the original document. Also, the SimCLS ranking uses intrinsic importance only (similarily with respect to the original document).

\paragraph{Divide-and-conquer approaches for long document summarization} The DANCER model of \newcite{gidiotis2020divide} breaks a summarization task into multiple smaller sequence-to-sequence sub-tasks, which share similar motivation to our intrinsic importance model. However, they use several heuristics to select specific sections of the papers (introduction, methods, results, and conclusion) while our sampling approach is domain agnostic. Recently, \citet{mao_dyle_2022} proposed DYLE, an \emph{extract-then-generate} method that extract text snippets from chunks of the input document, obtaining state-of-the-art results on GovReport. \citet{zhang_summn_2022} presented a multi-stage summarization method that generates coarse summaries for each document segment, which are then used to obtain a fine-grained summary. \citet{cao_hibrids_2022} add a learnable hierarchical bias term to the transformer attention mechanism, which allows the model to capture information about the structure (sections) of long documents.

\paragraph{Controllable summarization} 
Controlling length in summaries has been addressed by leveraging positional encodings \citep{takase_positional_2019}, a length-aware attention mechanism \citep{liu_length_2022}, and optimization objectives that include a length constraint \citep{makino_global_2019}. \citet{kikuchi_controlling_2016} explore different length control techniques at the learning and decoding stages. Control of other attributes such as entity coverage and summary style were achieved with control tokens \citep{fan2017controllable, he_ctrlsum_2020} and constrained Markov Decision Processes \citep{chan_controllable_2021}.

Similar to this work, GSum \citep{dou_gsum_2021} uses content guidance to improve summary quality. However, we note that GSum's guidance is used as \emph{input} of its sequence-to-sequence model, and shifts in guidance distribution would require further training. In contrast, FactorSum allows one to change the budget or content guidance without expensive retraining. Regarding evaluation, the best GSum variant achieves 45.09 ROUGE-1 on PubMed, whereas FactorSum achieves 45.41 R-1 without content guidance and 47.5 R-1 with content guidance. Finally, our sequence-to-sequence architecture is based on BART-base, thus requiring significantly fewer training parameters than GSum's dual BART-large encoders.

\section{Conclusion and Future Work}
In this work, we embrace the idea that general-purpose summary is an elusive goal and that contextual factors such as preferences for summary lengths are essential in the design of summarization systems \citep{jones1999automatic}. We propose a framework to separate budget decisions from selecting important content in the document using neural sequence-to-sequence models as building blocks. Our results suggest improved performance in both in-domain and cross-domain summarization of long documents. In future work, we plan to investigate the effects of domain-specific extrinsic guidance that would encourage the summaries to cover aspects of interest. % (e.g., methods, results, and conclusions in scientific papers).

\section*{Limitations}
\label{sec:limitations}
While our model requires fewer compute resources for training, the inference step is more expensive. For each document, $n_d=20$ document views are sampled and $n_d$ feedforward computations are performed for the intrinsic model (BART-base) before the greedy summary generation algorithm is applied. In our experiments, BART-base averages $0.27 \times n_d$ seconds per sample. The greedy generation adds in the worst case an average of 1.5 seconds per sample using a naive single-threaded implementation, which gives a total of 6.9 seconds per document. Fortunately, these computations are highly parallelizable, and more careful tuning of the number of views per document $n_d$ would make the runtime similar to a single large neural model. For comparison, PEGASUS and BigBird-PEGASUS take on average 3.13 and 3.85 seconds per sample on a single GPU (batch size = 4).

\section*{Ethical Considerations}
In Section \ref{sec:limitations}, we discussed how our model requires more compute power during inference, despite being relatively cheaper to train. Thus, if similar methods are used as part of a service in a production environment that serves a large number of requests, the inference procedure may result in unnecessary use of compute resources. In this case, we recommend careful tuning of the document sampling parameters $n_d$ and $s_f$ (Section \ref{sec:sampling_document_views}) to keep the number of summary views adequate for the appliceeds. In Appendix \ref{sec:document_sampling_details}, we provide some heuristics to optimize those sampling parameters.

\section*{Acknowledgements}
This work was supported by Actelligent Capital and used the Cirrus UK National Tier-2 HPC Service at EPCC (http://www.cirrus.ac.uk) funded by the University of Edinburgh and EPSRC (EP/P020267/1). We also thank Yifu Qiu and the anonymous reviewers for their insightful feedback.

% Entries for the entire Anthology, followed by custom entries
\bibliography{anthology,custom}

\begin{thebibliography}{33}
\expandafter\ifx\csname natexlab\endcsname\relax\def\natexlab#1{#1}\fi

\bibitem[{Beltagy et~al.(2020)Beltagy, Peters, and
  Cohan}]{beltagy2020longformer}
Iz~Beltagy, Matthew~E Peters, and Arman Cohan. 2020.
\newblock \href {https://arxiv.org/abs/2004.05150} {Longformer: The
  long-document transformer}.
\newblock \emph{ArXiv preprint}, abs/2004.05150.

\bibitem[{Cao and Wang(2022)}]{cao_hibrids_2022}
Shuyang Cao and Lu~Wang. 2022.
\newblock \href {https://doi.org/10.18653/v1/2022.acl-long.58} {{{HIBRIDS}}:
  {{Attention}} with {{Hierarchical Biases}} for {{Structure-aware Long
  Document Summarization}}}.
\newblock In \emph{Proceedings of the 60th {{Annual Meeting}} of the
  {{Association}} for {{Computational Linguistics}} ({{Volume}} 1: {{Long
  Papers}})}, pages 786--807, {Dublin, Ireland}. {Association for Computational
  Linguistics}.

\bibitem[{Chan et~al.(2021)Chan, Wang, and King}]{chan_controllable_2021}
Hou~Pong Chan, Lu~Wang, and Irwin King. 2021.
\newblock \href {https://doi.org/10.1162/tacl_a_00423} {Controllable
  summarization with constrained {M}arkov decision process}.
\newblock \emph{Transactions of the Association for Computational Linguistics},
  9:1213--1232.

\bibitem[{Cohan et~al.(2018)Cohan, Dernoncourt, Kim, Bui, Kim, Chang, and
  Goharian}]{cohan2018discourse}
Arman Cohan, Franck Dernoncourt, Doo~Soon Kim, Trung Bui, Seokhwan Kim, Walter
  Chang, and Nazli Goharian. 2018.
\newblock \href {https://doi.org/10.18653/v1/N18-2097} {A discourse-aware
  attention model for abstractive summarization of long documents}.
\newblock In \emph{Proceedings of the 2018 Conference of the North {A}merican
  Chapter of the Association for Computational Linguistics: Human Language
  Technologies, Volume 2 (Short Papers)}, pages 615--621, New Orleans,
  Louisiana. Association for Computational Linguistics.

\bibitem[{Dong et~al.(2021)Dong, Mircea, and Cheung}]{dong-etal-2021-discourse}
Yue Dong, Andrei Mircea, and Jackie Chi~Kit Cheung. 2021.
\newblock \href {https://doi.org/10.18653/v1/2021.eacl-main.93}
  {Discourse-aware unsupervised summarization for long scientific documents}.
\newblock In \emph{Proceedings of the 16th Conference of the European Chapter
  of the Association for Computational Linguistics: Main Volume}, pages
  1089--1102, Online. Association for Computational Linguistics.

\bibitem[{Dou et~al.(2021)Dou, Liu, Hayashi, Jiang, and Neubig}]{dou_gsum_2021}
Zi-Yi Dou, Pengfei Liu, Hiroaki Hayashi, Zhengbao Jiang, and Graham Neubig.
  2021.
\newblock \href {https://doi.org/10.18653/v1/2021.naacl-main.384} {{GS}um: A
  general framework for guided neural abstractive summarization}.
\newblock In \emph{Proceedings of the 2021 Conference of the North American
  Chapter of the Association for Computational Linguistics: Human Language
  Technologies}, pages 4830--4842, Online. Association for Computational
  Linguistics.

\bibitem[{Fan et~al.(2018)Fan, Grangier, and Auli}]{fan2017controllable}
Angela Fan, David Grangier, and Michael Auli. 2018.
\newblock \href {https://doi.org/10.18653/v1/W18-2706} {Controllable
  abstractive summarization}.
\newblock In \emph{Proceedings of the 2nd Workshop on Neural Machine
  Translation and Generation}, pages 45--54, Melbourne, Australia. Association
  for Computational Linguistics.

\bibitem[{Gidiotis and Tsoumakas(2020)}]{gidiotis2020divide}
Alexios Gidiotis and Grigorios Tsoumakas. 2020.
\newblock A divide-and-conquer approach to the summarization of long documents.
\newblock \emph{IEEE/ACM Transactions on Audio, Speech, and Language
  Processing}, 28:3029--3040.

\bibitem[{He et~al.(2020)He, Kry{\'s}ci{\'n}ski, McCann, Rajani, and
  Xiong}]{he_ctrlsum_2020}
Junxian He, Wojciech Kry{\'s}ci{\'n}ski, Bryan McCann, Nazneen Rajani, and
  Caiming Xiong. 2020.
\newblock \href {https://arxiv.org/abs/2012.04281} {{{CTRLsum}}: {{Towards
  Generic Controllable Text Summarization}}}.

\bibitem[{Hinton et~al.(2015)Hinton, Vinyals, and Dean}]{hinton2015distilling}
Geoffrey Hinton, Oriol Vinyals, and Jeff Dean. 2015.
\newblock \href {https://arxiv.org/abs/1503.02531} {Distilling the knowledge in
  a neural network}.
\newblock \emph{ArXiv preprint}, abs/1503.02531.

\bibitem[{Huang et~al.(2021)Huang, Cao, Parulian, Ji, and
  Wang}]{huang2021efficient}
Luyang Huang, Shuyang Cao, Nikolaus Parulian, Heng Ji, and Lu~Wang. 2021.
\newblock \href {https://doi.org/10.18653/v1/2021.naacl-main.112} {Efficient
  attentions for long document summarization}.
\newblock In \emph{Proceedings of the 2021 Conference of the North American
  Chapter of the Association for Computational Linguistics: Human Language
  Technologies}, pages 1419--1436, Online. Association for Computational
  Linguistics.

\bibitem[{Iso et~al.(2021)Iso, Wang, Suhara, Angelidis, and
  Tan}]{iso2021convex}
Hayate Iso, Xiaolan Wang, Yoshihiko Suhara, Stefanos Angelidis, and Wang-Chiew
  Tan. 2021.
\newblock \href {https://doi.org/10.18653/v1/2021.findings-emnlp.328} {{C}onvex
  {A}ggregation for {O}pinion {S}ummarization}.
\newblock In \emph{Findings of the Association for Computational Linguistics:
  EMNLP 2021}, pages 3885--3903, Punta Cana, Dominican Republic. Association
  for Computational Linguistics.

\bibitem[{Jones et~al.(1999)}]{jones1999automatic}
K~Sparck Jones et~al. 1999.
\newblock Automatic summarizing: factors and directions.
\newblock \emph{Advances in automatic text summarization}, pages 1--12.

\bibitem[{Kikuchi et~al.(2016)Kikuchi, Neubig, Sasano, Takamura, and
  Okumura}]{kikuchi_controlling_2016}
Yuta Kikuchi, Graham Neubig, Ryohei Sasano, Hiroya Takamura, and Manabu
  Okumura. 2016.
\newblock \href {https://doi.org/10.18653/v1/D16-1140} {Controlling output
  length in neural encoder-decoders}.
\newblock In \emph{Proceedings of the 2016 Conference on Empirical Methods in
  Natural Language Processing}, pages 1328--1338, Austin, Texas. Association
  for Computational Linguistics.

\bibitem[{LeCun et~al.(2006)LeCun, Chopra, Hadsell, Ranzato, and
  Huang}]{lecun2006tutorial}
Yann LeCun, Sumit Chopra, Raia Hadsell, M~Ranzato, and F~Huang. 2006.
\newblock A tutorial on energy-based learning.
\newblock \emph{Predicting structured data}, 1(0).

\bibitem[{Levenshtein et~al.(1966)}]{levenshtein1966binary}
Vladimir~I Levenshtein et~al. 1966.
\newblock Binary codes capable of correcting deletions, insertions, and
  reversals.
\newblock In \emph{Soviet physics doklady}, volume~10, pages 707--710. Soviet
  Union.

\bibitem[{Lewis et~al.(2020)Lewis, Liu, Goyal, Ghazvininejad, Mohamed, Levy,
  Stoyanov, and Zettlemoyer}]{lewis2019bart}
Mike Lewis, Yinhan Liu, Naman Goyal, Marjan Ghazvininejad, Abdelrahman Mohamed,
  Omer Levy, Veselin Stoyanov, and Luke Zettlemoyer. 2020.
\newblock \href {https://doi.org/10.18653/v1/2020.acl-main.703} {{BART}:
  Denoising sequence-to-sequence pre-training for natural language generation,
  translation, and comprehension}.
\newblock In \emph{Proceedings of the 58th Annual Meeting of the Association
  for Computational Linguistics}, pages 7871--7880, Online. Association for
  Computational Linguistics.

\bibitem[{Lin(2004)}]{lin2004rouge}
Chin-Yew Lin. 2004.
\newblock \href {https://aclanthology.org/W04-1013} {{ROUGE}: A package for
  automatic evaluation of summaries}.
\newblock In \emph{Text Summarization Branches Out}, pages 74--81, Barcelona,
  Spain. Association for Computational Linguistics.

\bibitem[{Liu and Liu(2021)}]{liu2021simcls}
Yixin Liu and Pengfei Liu. 2021.
\newblock \href {https://doi.org/10.18653/v1/2021.acl-short.135} {{S}im{CLS}: A
  simple framework for contrastive learning of abstractive summarization}.
\newblock In \emph{Proceedings of the 59th Annual Meeting of the Association
  for Computational Linguistics and the 11th International Joint Conference on
  Natural Language Processing (Volume 2: Short Papers)}, pages 1065--1072,
  Online. Association for Computational Linguistics.

\bibitem[{Liu et~al.(2022)Liu, Jia, and Zhu}]{liu_length_2022}
Yizhu Liu, Qi~Jia, and Kenny Zhu. 2022.
\newblock \href {https://doi.org/10.18653/v1/2022.acl-long.474} {Length
  {{Control}} in {{Abstractive Summarization}} by {{Pretraining Information
  Selection}}}.
\newblock In \emph{Proceedings of the 60th {{Annual Meeting}} of the
  {{Association}} for {{Computational Linguistics}} ({{Volume}} 1: {{Long
  Papers}})}, pages 6885--6895, {Dublin, Ireland}. {Association for
  Computational Linguistics}.

\bibitem[{Makino et~al.(2019)Makino, Iwakura, Takamura, and
  Okumura}]{makino_global_2019}
Takuya Makino, Tomoya Iwakura, Hiroya Takamura, and Manabu Okumura. 2019.
\newblock \href {https://doi.org/10.18653/v1/P19-1099} {Global optimization
  under length constraint for neural text summarization}.
\newblock In \emph{Proceedings of the 57th Annual Meeting of the Association
  for Computational Linguistics}, pages 1039--1048, Florence, Italy.
  Association for Computational Linguistics.

\bibitem[{Mao et~al.(2021)Mao, Wu, Ni, Zhang, Zhang, Yu, Deb, Zhu, Awadallah,
  and Radev}]{mao_dyle_2022}
Ziming Mao, Chen~Henry Wu, Ansong Ni, Yusen Zhang, Rui Zhang, Tao Yu,
  Budhaditya Deb, Chenguang Zhu, Ahmed~H. Awadallah, and Dragomir Radev. 2021.
\newblock \href {https://arxiv.org/abs/2110.08168} {{{DYLE}}: {{Dynamic Latent
  Extraction}} for {{Abstractive Long-Input Summarization}}}.

\bibitem[{Mihalcea and Tarau(2004)}]{mihalcea2004textrank}
Rada Mihalcea and Paul Tarau. 2004.
\newblock \href {https://aclanthology.org/W04-3252} {{T}ext{R}ank: Bringing
  order into text}.
\newblock In \emph{Proceedings of the 2004 Conference on Empirical Methods in
  Natural Language Processing}, pages 404--411, Barcelona, Spain. Association
  for Computational Linguistics.

\bibitem[{Oved and Levy(2021)}]{oved2021pass}
Nadav Oved and Ran Levy. 2021.
\newblock \href {https://doi.org/10.18653/v1/2021.acl-long.30} {{PASS}:
  Perturb-and-select summarizer for product reviews}.
\newblock In \emph{Proceedings of the 59th Annual Meeting of the Association
  for Computational Linguistics and the 11th International Joint Conference on
  Natural Language Processing (Volume 1: Long Papers)}, pages 351--365, Online.
  Association for Computational Linguistics.

\bibitem[{Peyrard(2019)}]{peyrard2018simple}
Maxime Peyrard. 2019.
\newblock \href {https://doi.org/10.18653/v1/P19-1101} {A simple theoretical
  model of importance for summarization}.
\newblock In \emph{Proceedings of the 57th Annual Meeting of the Association
  for Computational Linguistics}, pages 1059--1073, Florence, Italy.
  Association for Computational Linguistics.

\bibitem[{Rajbhandari et~al.(2019)Rajbhandari, Rasley, Ruwase, and
  He}]{rajbhandari2019zero}
Samyam Rajbhandari, Jeff Rasley, Olatunji Ruwase, and Yuxiong He. 2019.
\newblock \href
  {https://www.microsoft.com/en-us/research/publication/zero-memory-optimizations-toward-training-trillion-parameter-models/}
  {Zero: Memory optimizations toward training trillion parameter models}.
\newblock ArXiv.

\bibitem[{Sun et~al.(2019)Sun, Shapira, Dagan, and Nenkova}]{sun2019compare}
Simeng Sun, Ori Shapira, Ido Dagan, and Ani Nenkova. 2019.
\newblock \href {https://doi.org/10.18653/v1/W19-2303} {How to compare
  summarizers without target length? pitfalls, solutions and re-examination of
  the neural summarization literature}.
\newblock In \emph{Proceedings of the Workshop on Methods for Optimizing and
  Evaluating Neural Language Generation}, pages 21--29, Minneapolis, Minnesota.
  Association for Computational Linguistics.

\bibitem[{Sutskever et~al.(2014)Sutskever, Vinyals, and
  Le}]{sutskever2014sequence}
Ilya Sutskever, Oriol Vinyals, and Quoc~V. Le. 2014.
\newblock \href
  {https://proceedings.neurips.cc/paper/2014/hash/a14ac55a4f27472c5d894ec1c3c743d2-Abstract.html}
  {Sequence to sequence learning with neural networks}.
\newblock In \emph{Advances in Neural Information Processing Systems 27: Annual
  Conference on Neural Information Processing Systems 2014, December 8-13 2014,
  Montreal, Quebec, Canada}, pages 3104--3112.

\bibitem[{Takase and Okazaki(2019)}]{takase_positional_2019}
Sho Takase and Naoaki Okazaki. 2019.
\newblock \href {https://doi.org/10.18653/v1/N19-1401} {Positional encoding to
  control output sequence length}.
\newblock In \emph{Proceedings of the 2019 Conference of the North {A}merican
  Chapter of the Association for Computational Linguistics: Human Language
  Technologies, Volume 1 (Long and Short Papers)}, pages 3999--4004,
  Minneapolis, Minnesota. Association for Computational Linguistics.

\bibitem[{Vijayakumar et~al.(2016)Vijayakumar, Cogswell, Selvaraju, Sun, Lee,
  Crandall, and Batra}]{vijayakumar2016diverse}
Ashwin~K Vijayakumar, Michael Cogswell, Ramprasath~R Selvaraju, Qing Sun,
  Stefan Lee, David Crandall, and Dhruv Batra. 2016.
\newblock \href {https://arxiv.org/abs/1610.02424} {Diverse beam search:
  Decoding diverse solutions from neural sequence models}.
\newblock \emph{ArXiv preprint}, abs/1610.02424.

\bibitem[{Zaheer et~al.(2020)Zaheer, Guruganesh, Dubey, Ainslie, Alberti,
  Onta{\~{n}}{\'{o}}n, Pham, Ravula, Wang, Yang, and Ahmed}]{zaheer2020big}
Manzil Zaheer, Guru Guruganesh, Kumar~Avinava Dubey, Joshua Ainslie, Chris
  Alberti, Santiago Onta{\~{n}}{\'{o}}n, Philip Pham, Anirudh Ravula, Qifan
  Wang, Li~Yang, and Amr Ahmed. 2020.
\newblock \href
  {https://proceedings.neurips.cc/paper/2020/hash/c8512d142a2d849725f31a9a7a361ab9-Abstract.html}
  {Big bird: Transformers for longer sequences}.
\newblock In \emph{Advances in Neural Information Processing Systems 33: Annual
  Conference on Neural Information Processing Systems 2020, NeurIPS 2020,
  December 6-12, 2020, virtual}.

\bibitem[{Zhang et~al.(2020)Zhang, Zhao, Saleh, and Liu}]{zhang2020pegasus}
Jingqing Zhang, Yao Zhao, Mohammad Saleh, and Peter~J. Liu. 2020.
\newblock \href {http://proceedings.mlr.press/v119/zhang20ae.html} {{PEGASUS:}
  pre-training with extracted gap-sentences for abstractive summarization}.
\newblock In \emph{Proceedings of the 37th International Conference on Machine
  Learning, {ICML} 2020, 13-18 July 2020, Virtual Event}, volume 119 of
  \emph{Proceedings of Machine Learning Research}, pages 11328--11339. {PMLR}.

\bibitem[{Zhang et~al.(2021)Zhang, Ni, Mao, Wu, Zhu, Deb, Awadallah, Radev, and
  Zhang}]{zhang_summn_2022}
Yusen Zhang, Ansong Ni, Ziming Mao, Chen~Henry Wu, Chenguang Zhu, Budhaditya
  Deb, Ahmed~H. Awadallah, Dragomir Radev, and Rui Zhang. 2021.
\newblock \href {https://arxiv.org/abs/2110.10150} {Summ\^{{N}}: {{A
  Multi-Stage Summarization Framework}} for {{Long Input Dialogues}} and
  {{Documents}}}.

\end{thebibliography}
\bibliographystyle{acl_natbib}
\clearpage
\appendix

\section{Document Sampling Experiments}
\label{sec:document_sampling_details}
The sampling factor ($s_f$) and number of samples per document ($n_d$) are important hyperparameters that affect the summarization performance and computation costs of \textsc{\modelname}. By increasing the number of samples per document, the coverage of oracle sentences is also increased. Also, a smaller sampling factor will make each document view shorter and less prone to input truncation, at the cost of oracle sentence coverage. 

For our experimental setup in Sections \ref{sec:experiments} and \ref{sec:results}, we pick the sampling factor so that the number of sentences/tokens fit BART input limit (1024 tokens) with minimal truncation. The number of samples per document is chosen so that the coverage of oracle sentences in the original article is close to 100\%, while keeping the resulting dataset size and training costs manageable. According to Table \ref{tab:document_sampling}, a sampling factor $s_f=0.2$ and number of samples $n_d=20$ fullfil the requirements above for all datasets.

\begin{table}[H]
  \centering
  \setlength\tabcolsep{2.pt}
  \begin{tabular}{cccccc}
    \toprule
    \multicolumn{1}{c}{\textbf{Sampling}} & \multicolumn{3}{c}{\textbf{Oracle coverage (\%)}} & \textbf{Average} \\
    %\cmidrule(r){2-3}
    \multicolumn{1}{c}{\textbf{factor ($s_f$)}} & $n_d=5$ & $n_d=10$ & $n_d=20$ & \textbf{sentences} \\
    \midrule
    \multicolumn{5}{c}{\textbf{PubMed}} \\
    \midrule
    $0.1$ & 38.9 & 65.1 & 87.6 & 8.4  \\
    $0.2$ & 66.7 & 90.2 & 99.1 & 17.2  \\
    $0.25$ & 77.0 & 95.6 & 100 & 21.6  \\
    $0.33$ & 87.9 & 99.7 & 100 & 29.0 \\
    $0.5$ & 98.3 & 100 & 100 & 43.8 \\
    \midrule
    \multicolumn{5}{c}{\textbf{arXiv}} \\
    \midrule
    $0.1$ & 37.9 & 63.8 & 88.1 & 27.6  \\
    $0.2$ & 64.6 & 88.4 & 99.0 & 55.7  \\
    $0.25$ & 75.6 & 93.3 & 99.7 & 69.8  \\
    $0.33$ & 86.8 & 97.7 & 100 & 93.2 \\
    $0.5$ & 96.5 & 99.9 & 100 & 140.0 \\
    \midrule
    \multicolumn{5}{c}{\textbf{GovReport}} \\
    \midrule
    $0.1$ & 39.6 & 65.2 & 88.5 & 29.8  \\
    $0.2$ & 65.4 & 88.7 & 99.0 & 60.2  \\
    $0.25$ & 76.0 & 93.4 & 99.7 & 75.4  \\
    $0.33$ & 86.4 & 97.5 & 100 & 100.7 \\
    $0.5$ & 96.6 & 100 & 100 & 151.3 \\
    \bottomrule
  \end{tabular}
  \caption{Oracle sentence coverage and average number of sentences in sampled documents (validation sets) for different configurations of sampling factor $s_f$ and samples per document $n_d$.}\label{tab:document_sampling}
\end{table}

\begin{figure}
    \centering
    \includegraphics[width=0.5\textwidth]{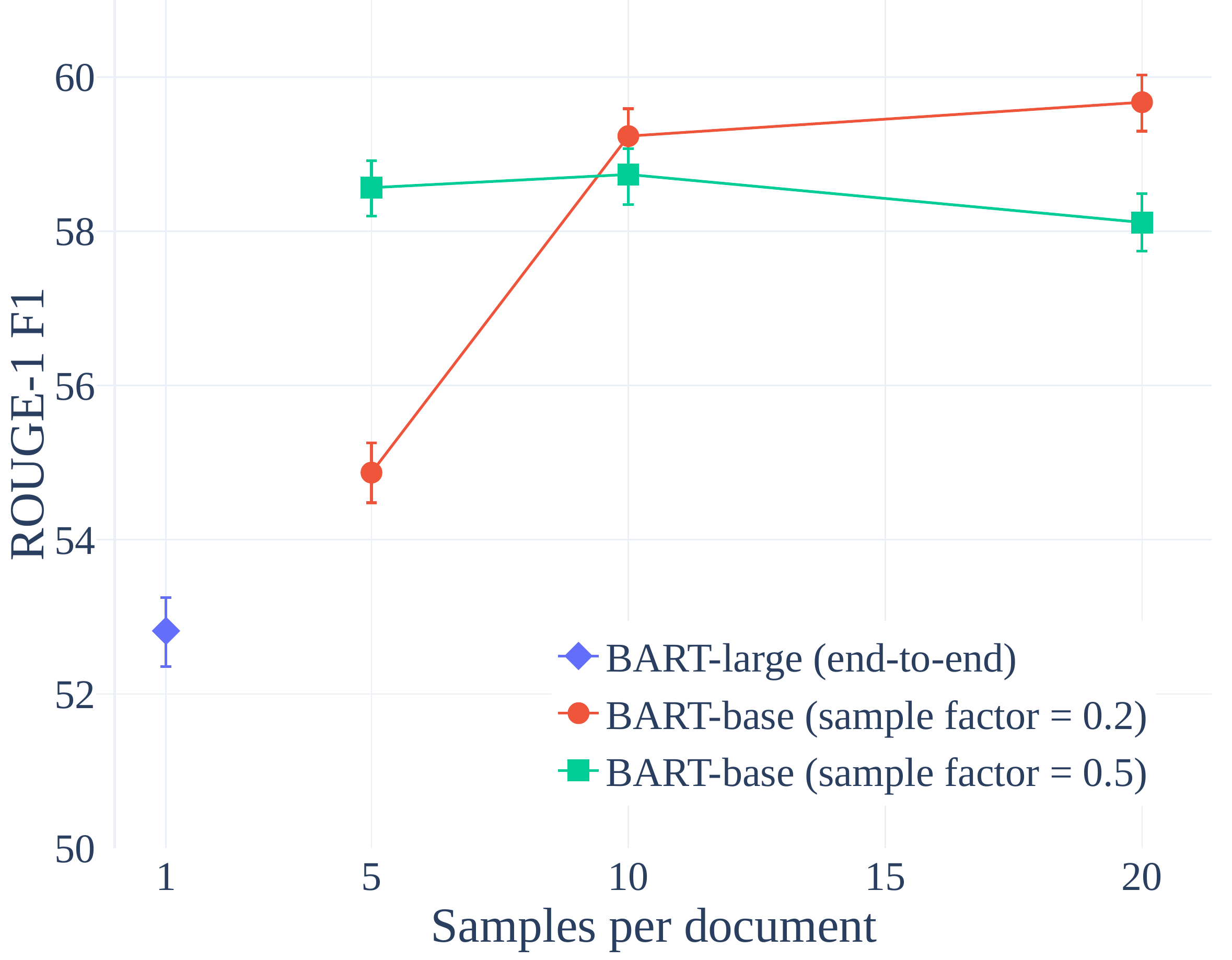}
    \caption{ROUGE-1 (F1) scores for different values of sampling factor ($s_f$) and number of samples per document ($n_d$), evaluated on the GovReport test set. BART-large is an end-to-end baseline, which is equivalent to $n_d=1$ and $s_f=1$.}\label{fig:sample_factors}
\end{figure}

To further investigate the effects of different sampling strategies, we train \textsc{\modelname} versions with  sampling factor $s_f \in \{0.5, 0.2\}$ and samples per document $n_d \in \{5, 10, 20\}$. In Figure \ref{fig:sample_factors}, we report evaluation results on the GovReport test set, which confirm the importance of oracle sentence coverage for the final summarization performance. Specifically, for a fixed $n_d=5$, the model with $s_f=0.2$ (65.4\% oracle coverage) achieves 54.87 ROUGE-1 versus 58.57 ROUGE-1 for the version $s_f=0.5$ (96.6\% oracle coverage). A model with the same sampling factor $s_f=0.2$ but using more samples per document achieves up to 59.67 ROUGE-1 ($n_d=5$, 99.1\% oracle coverage).

Finally, we observe that lower sampling factors achieve higher ROUGE scores, specially compared to the BART-large end-to-end baseline ($s_f=1$, $n_d=1$). These results suggest that working with shorter inputs is beneficial for summarization. We believe that this difference in performance is due to less truncation of the inputs. 

\section{Evaluation Results for Varying Budgets}
\begin{figure}
    \centering
    \begin{subfigure}[b]{.49\textwidth}
         \centering
         \includegraphics[width=\textwidth]{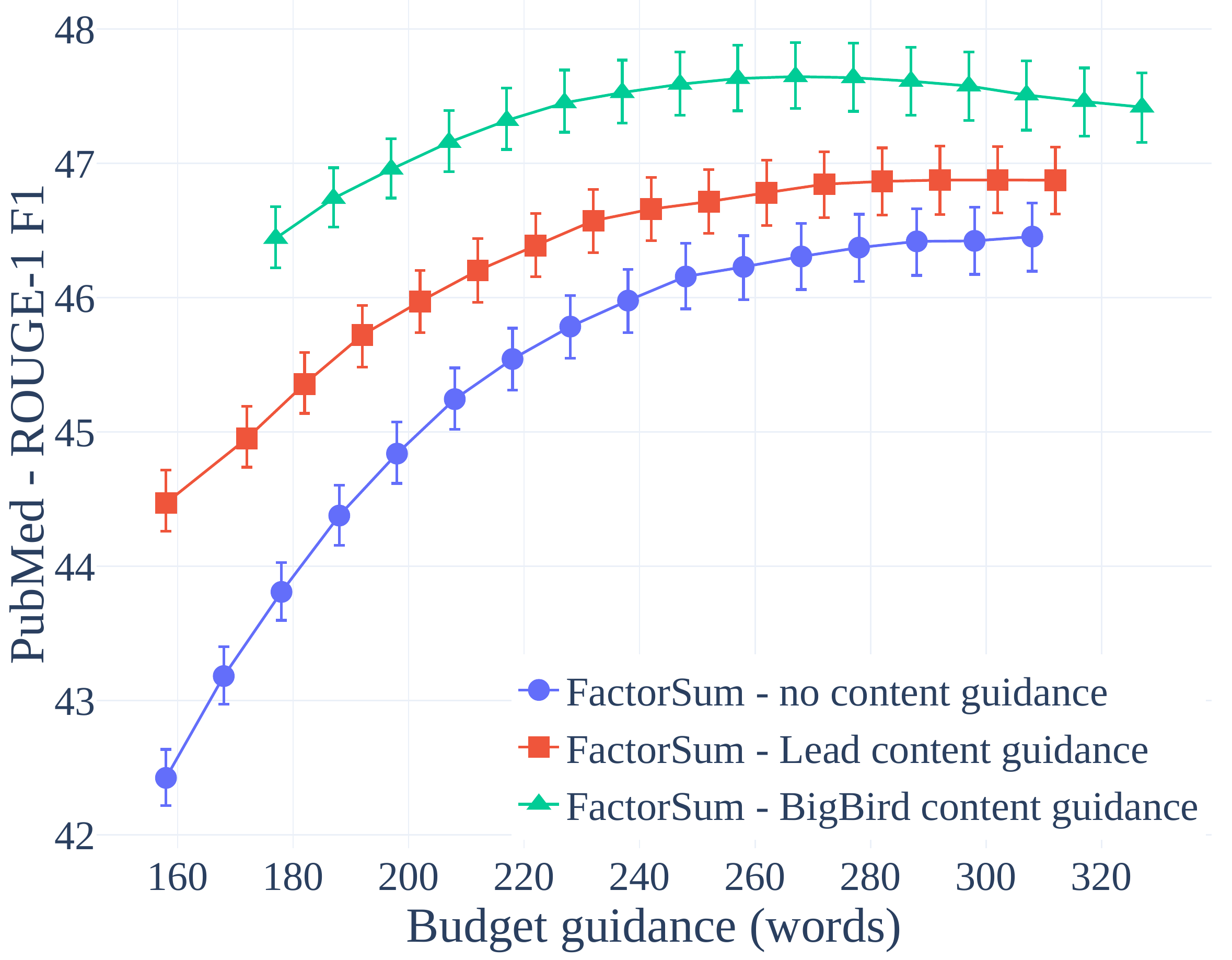}
         %\caption{}
         \label{fig:budgets_fmeasure_pubmed}
     \end{subfigure}
     \hfill
     \begin{subfigure}[b]{.49\textwidth}
         \centering
         \includegraphics[trim={0cm 0cm 0cm 0cm},clip,width=\textwidth]{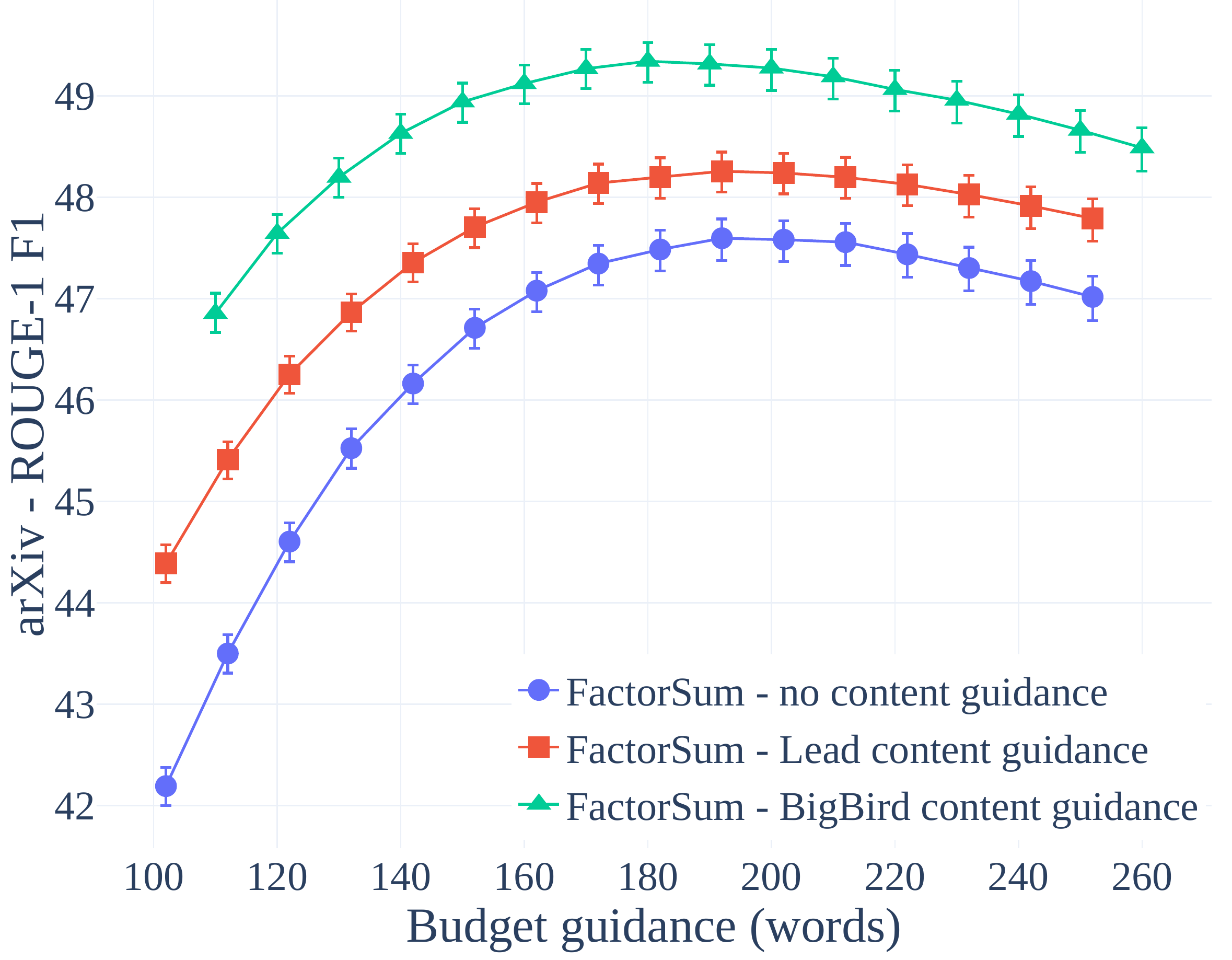}
         %\caption{}
         \label{fig:budgets_fmeasure_arxiv}
     \end{subfigure}
     \begin{subfigure}[b]{.49\textwidth}
         \centering
         \includegraphics[trim={0cm 0cm 0cm 0cm},clip,width=\textwidth]{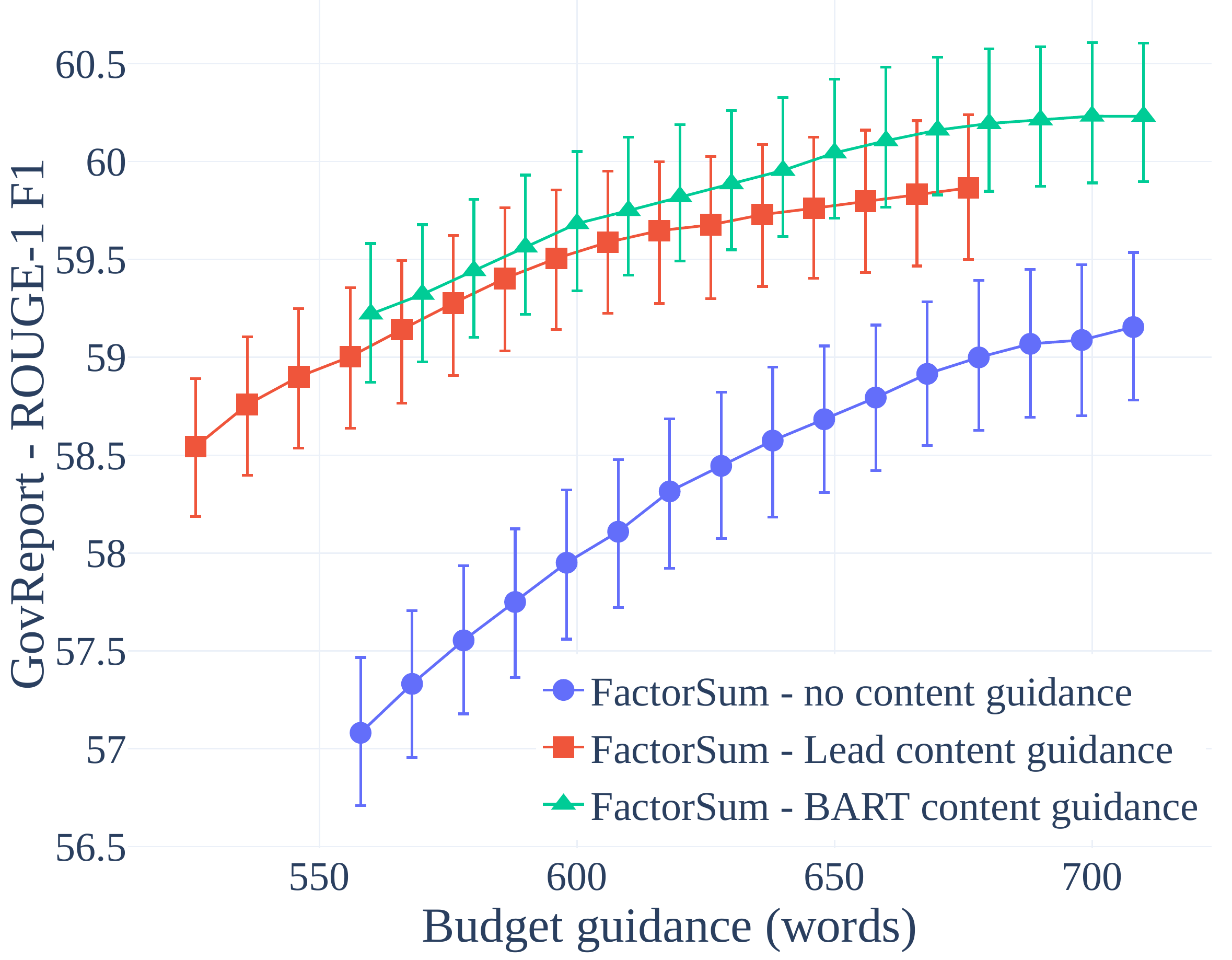}
         \label{fig:budgets_fmeasure_govreport}
     \end{subfigure}
    \vspace{-1.2\baselineskip}
    \caption{ROUGE-1 (F1) scores for different summary budget and content guidance computed on PubMed (top), arXiv (middle), and GovReport (bottom) test sets. Error bars indicate 95\% confidence interval.}
    \label{fig:budgets_fmeasure}
\end{figure}

In Figure \ref{fig:budgets_fmeasure}, we provide results for varying budget guidance values on PubMed, arXiv, and GovReport test sets. For all datasets, there is a consistent improvement for content-guided summaries versus \textsc{\modelname} without content guidance. We also note that BigBird content guidance leads to significant improvements on PubMed and arXiv but BART-large is statistically equivalent to Lead guidance on GovReport, which means there is still room for improvement in our end-to-end BART-large baseline. 

\label{sec:budgets_fmeasure}

\section{Training Details}
\label{sec:training_details}
The intrinsic importance model $p_\theta(R^{(i)}_v | D^{(i)}_v)$ described in Section \ref{sec:intrinsic_importance} is implemented using the BART sequence-to-sequence model \citep{lewis2019bart}. We fine-tune the \texttt{bart-base} checkpoint from huggingface\footnote{\url{https://huggingface.co/facebook/bart-base}} on the datasets of document and summary views $\mathcal{T'}$ presented in Section \ref{sec:sampling_document_views}. Unless otherwise stated, we use $n_d=20$ samples per document and a sampling factor $s_f=0.2$, as explained in Appendix \ref{sec:document_sampling_details}. To ensure replicability, we use a random seed for document views sampling.

For the training process, we use 4 GeForce GTX 1080 Ti GPUs each with 12GB of memory. Table \ref{tab:training_details} details  the training set size (number of documents and summary views), number of training steps, and time to train the intrinsic importance models for each dataset. The main training hyperparameters are presented in Table \ref{tab:training_parameters}.

\begin{table}
  \centering
  \setlength\tabcolsep{2.9pt}
  \begin{tabular}{lcccc}
    \toprule
    \textbf{Dataset} & \textbf{Docs} & \textbf{\makecell{Summary\\Views}} & \textbf{Steps} & \textbf{Hours} \\ 
    \midrule
    PubMed & 115k & 2.3M & 100k & 91  \\
    arXiv & 200k & 4M & 200k & 180  \\
    GovReport & 17.5k & 340k & 50k & 51  \\
    \bottomrule
  \end{tabular}
  \caption{Statistics for number of documents, summary views, training steps, and total training time for each dataset (intrinsic model based on BART-base).}\label{tab:training_details}
\end{table}

\begin{table}
  \centering
  \setlength\tabcolsep{2.9pt}
  \begin{tabular}{lcc}
    \toprule
    \textbf{Parameter} & \textbf{BART} & \textbf{BART} \\
     & \textbf{base} & \textbf{large} \\
    \midrule
    Checkpoint & base & large \\
    Number of parameters & 139.4M & 406.3M \\
    Training time & Table \ref{tab:training_details} & 224 hours \\
    Batch size per GPU & 4 & 1 \\
    Gradient accumulation & 4 & 32 \\
    Effective batch size & 64 & 32 \\
    Learning rate & $5 \times 10^{-5}$ & $2 \times 10^{-5}$ \\
    Learning rate scheduler & \multicolumn{2}{c}{linear} \\
    Optimizer & \multicolumn{2}{c}{AdamW} \\
    Adam $\beta_1$ & \multicolumn{2}{c}{0.9} \\
    Adam $\beta_2$ & \multicolumn{2}{c}{0.999} \\
    Adam $\epsilon$ & \multicolumn{2}{c}{$1 \times 10^{-8}$} \\
    Metric best model & \multicolumn{2}{c}{ROUGE-1 F1} \\
    Floating point precision & \multicolumn{2}{c}{FP16} \\
    Max source length & \multicolumn{2}{c}{1024} \\
    Max target length & 128 & 768 \\
    Generation beams & 4 & 8 \\
    Length penalty & 1.0 & 0.6 \\  
    DeepSpeed ZeRO\footnote{\url{https://www.deepspeed.ai/tutorials/zero}} & - & Stage 1 \\
    \bottomrule
  \end{tabular}
  \caption{Training details and  hyperparameters for the intrinsic model (BART-base) and the end-to-end baseline for GovReport (BART-large).}
  \label{tab:training_parameters}
\end{table}

\paragraph{End-to-end summarization}
We train our own end-to-end BART-large baseline on the GovReport dataset. Since the target summaries are long, we the maximum summary generation length to 768 tokens, which makes the memory requirements to exceed most single-GPU capacities (even with batch size equal to one). To address this problem, we resort to model parallelism techniques provided by the DeepSpeed library \cite{rajbhandari2019zero}, allowing the efficient distribution of the model across 4 GeForce GTX 1080 Ti GPUs, each with 12GB of memory. We use gradient accumulation to achieve a effective batch size of 32. Table \ref{tab:training_parameters} details the training hyperparameters.

\section{Inference Details}
\label{sec:inference_details}
For reproducibility purposes, we provide the generation details for the end-to-end baseline models in Table \ref{tab:baseline_generation_parameters}. The intrinsic importance model (BART-base) generation uses the same maximum source length, maximum target length, beam size, and length penalty defined for training in Table \ref{tab:training_parameters}.

\begin{table}
  \centering
  \setlength\tabcolsep{2.9pt}
  \begin{tabular}{lc}
    \toprule
    \multicolumn{2}{c}{\textbf{PEGASUS}} \\
    \midrule
    Checkpoint (arXiv) & pegasus-arxiv\\ %\tablefootnote{\url{https://huggingface.co/google/pegasus-arxiv}} \\
    Checkpoint (PubMed) & pegasus-pubmed\\ %\tablefootnote{\url{https://huggingface.co/google/pegasus-pubmed}} \\
    Number of parameters & 570.8M \\
    Max source length & 1024 \\
    Generation beams & 8 \\
    Length penalty & 0.8 \\  
    \midrule
    \multicolumn{2}{c}{\textbf{BigBird}} \\
    \midrule
    Checkpoint (arXiv) & bigbird-pegasus-\\
    & large-arxiv\\ %\tablefootnote{\url{https://huggingface.co/google/bigbird-pegasus-large-arxiv}}\\ 
    Checkpoint (PubMed) & bigbird-pegasus-\\
    & large-pubmed\\ %\tablefootnote{\url{https://huggingface.co/google/bigbird-pegasus-large-pubmed}}\\ 
    Number of parameters & 576.9M \\
    Max source length & 3072 \\
    Generation beams & 5 \\
    Length penalty & 0.8 \\
    \midrule
    \multicolumn{2}{c}{\textbf{BART-large}} \\
    \midrule
    Checkpoint (GovReport) & See Appendix \ref{sec:training_details} \\ 
    Number of parameters & 406.3M \\
    Max source length & 1024 \\
    Generation beams & 8 \\
    Length penalty & 1.0 \\
    \midrule
    \multicolumn{2}{c}{\textbf{All models}} \\
    \midrule
    Max target length & 256 \\
    (arXiv, PubMed) & \\
    Max target length & 768 \\
    (GovReport) & \\
    \bottomrule
  \end{tabular}
  \caption{Summary generation details and parameters for the end-to-end baselines.}\label{tab:baseline_generation_parameters}
\end{table}

\paragraph{Extrinsic importance model}
For summary length control, we adjust the budget guidance so that the average summary lengths is close to the average length of the first 1,000 summaries from the validation set (or the entire validation set for GovReport). The budget guidance for each model/guidance type is shown in Table \ref{tab:budget_guidance}. For the domain adaptation results in Table \ref{tab:domain_adaptation}, the budget guidance corresponds to the target dataset, i.e., it is the in-domain budget listed in Table \ref{tab:budget_guidance}.

\begin{table}
  \centering
  \setlength\tabcolsep{2.2pt}
  \begin{tabular}{cc|c|c|c}
    \toprule
    \multicolumn{2}{c|}{\textbf{Guidance}} & \multirow{2}{1.4cm}{\centering \textbf{Pubmed}} & \multirow{2}{1.cm}{\centering \textbf{arXiv}} & \multirow{2}{1.9cm}{\centering \textbf{GovReport}} \\
    \textbf{Budget} & \textbf{Content} & \, & \\
    \toprule
    \multicolumn{5}{c}{\textbf{\textsc{\modelname} - no content guidance}} \\
    \midrule
    Oracle & - & 216 & 169 & 656 \\
    Fixed & - & 213 & 167 & 656 \\
    Model & - & 217 & 170 & 698 \\
    \midrule
    \multicolumn{5}{c}{\textbf{\textsc{\modelname} - content guidance}} \\
    \midrule
    Oracle & Lead & 221 & 169 & 632 \\
    Fixed & Lead & 217 & 167 & 624 \\
    Fixed & Model & 232 & 175 & 658 \\
    Model & Model & 227 & 177 & 658 \\
    \bottomrule
  \end{tabular}
  \caption{Budget guidance used for \textsc{\modelname} models in Table \ref{tab:experiment_results}. Model guidance is provided by BART-large for GovReport and BigBird for PubMed and arXiv.}\label{tab:budget_guidance}
\end{table}

In addition to the inference procedure described in Algorithm \ref{alg:greedy_summary}, we apply a preprocessing step that divides each summary view $S_v \in V_D$ into sentences. The resulting sentence-tokenized set of summary views $V_D$ is the union of all sentences. We use the sentence tokenizer provided by \texttt{NLTK}\footnote{\url{nltk.org}}. 

For \textsc{\modelname} versions using content guidance, the best summary $S^*$ returned by Algorithm \ref{alg:greedy_summary} is reordered according to the following procedure: (1) for each summary view in $S^*$, we collect the index of the oracle sentence in the content guidance text\footnote{Oracle sentences determined as described in Section \ref{sec:sampling_document_views}.}; (2) The summary views are sorted according to the list of corresponding oracle indexes, using the Python \texttt{sorted} function\footnote{\url{https://docs.python.org/3/library/functions.html\#sorted}}.

\section{Validation results}
\label{sec:validation_results}

In Table \ref{tab:experiment_results_validation}, we provide validation scores corresponding to the in-domain summarization test results in Table \ref{tab:experiment_results}. Additionally, Table \ref{tab:ablation_study_validation} shows the validation scores for the ensemble experiments corresponding to the test results in Table \ref{tab:ablation_study}. 

\begin{table}
  \centering
  \setlength\tabcolsep{1.5pt}
  \begin{tabular}{l|ccc|ccc}
    \toprule
    \multicolumn{1}{c|}{\multirow{2}{1.2cm}{\centering \textbf{Ranker}}} &  \multicolumn{3}{c|}{\textbf{PubMed}} & \multicolumn{3}{c}{\textbf{arXiv}} \\
    \cmidrule(r){2-4} \cmidrule(r){5-7} & \textbf{R-1} & \textbf{R-2} & \textbf{R-L} & \textbf{R-1} & \textbf{R-2} & \textbf{R-L} \\
    \toprule
    \multicolumn{7}{c}{\textbf{PEGASUS + BigBird Ensemble Summaries}} \\
    \midrule
    TextRank & 44.02 & 18.40 & 38.73 & 44.20 & 16.97 & 37.66 \\
    \midrule
    FS & 44.83 & \underline{\textbf{19.12}} & 40.90 & 44.77 & 17.36 & 39.84 \\
    FS-Oracle & 48.77 & 21.72 & 44.40 & 49.18 & 20.07 & 43.72 \\
    \midrule
    \multicolumn{7}{c}{\textbf{Summary Views}} \\
    \midrule
    TextRank & 42.17 & 16.82 & 37.62 & 42.49 & 16.39 & 37.58 \\
    \midrule
    FS & \underline{\textbf{45.33}} & 18.69 & \underline{\textbf{41.62}} & \underline{\textbf{47.16}} & \underline{\textbf{18.57}} & \underline{\textbf{42.57}} \\
    FS-Oracle & 51.64 & 23.27 & 47.48 & 53.27 & 22.75 & 48.08 \\
    \bottomrule
  \end{tabular}
  \caption{ROUGE F1 scores on the \textbf{validation sets} for the ensemble experiments. We compare summary predictions given by the concatenation of PEGASUS and BigBird summaries against summaries derived from \textsc{\modelname} summary views. We use two sentence rankers: an unsupervised TextRank baseline and \textsc{\modelname} extrinsic importance ranker. FS and FS-oracle use \textsc{\modelname} without content guidance and with reference summary guidance, respectively. All models use \emph{fixed budget} as described in Section \ref{sec:results-budget-guidance}. Best non-oracle results are \textbf{bold-faced}. \underline{Underlined results} are statistically equivalent to the best scores ($p<0.05$).}\label{tab:ablation_study_validation}
\end{table}

\begin{table*}[h]
  \centering
  \setlength\tabcolsep{3.5pt}
  \begin{tabular}{lc|cccc|cccc|cccc}
    \toprule
    \multicolumn{2}{c|}{\multirow{2}{2.2cm}{\centering \textbf{Model}}} & \multicolumn{4}{c|}{\textbf{PubMed}} & \multicolumn{4}{c|}{\textbf{arXiv}} & \multicolumn{4}{c}{\textbf{GovReport}} \\
    \cmidrule(r){3-6} \cmidrule(r){7-10} \cmidrule(r){11-14} 
    &  & \textbf{R-1} & \textbf{R-2} & \textbf{R-L} & \textbf{Len} & \textbf{R-1} & \textbf{R-2} & \textbf{R-L} & \textbf{Len} & \textbf{R-1} & \textbf{R-2} & \textbf{R-L} & \textbf{Len} \\
    \toprule
    \multicolumn{14}{c}{\textbf{Previous work}} \\
    \midrule
    \multicolumn{2}{l|}{PEGASUS} & 43.73 & 18.77 & 40.15 & 181 & 43.07 & 16.39 & 38.66 & 170 & - & - & - & - \\
    \multicolumn{2}{l|}{BigBird} & 45.28 & 19.77 & 41.60 & 186 & 46.02 & 18.54 & 41.35 & 164 & - & - & - & - \\
    \multicolumn{2}{l|}{BART-large} & - & - & - & - & - & - & - & - & 53.06 & 19.11 & 50.12 & 597 \\
    \midrule
    \multicolumn{2}{c|}{\textbf{Guidance}} & \multicolumn{12}{c}{} \\
    \cmidrule(r){1-2}
    \textbf{Budget} & \textbf{Content} & \multicolumn{9}{c}{\textbf{\textsc{\modelname} - no content guidance}} \\
    \midrule
    Oracle & - & 47.37 & 19.16 & 43.33 & 210 & 48.85 & 18.85 & 43.92 & 165 & 59.54 & 23.98 & 55.82 & 642 \\
    Fixed & - & 45.33 & 18.69 & 41.62 & 205 & 47.16 & 18.57 & 42.57 & 165 & 58.41 & 23.90 & 54.83 & 650 \\
    Model & - & 44.71 & 18.07 & 40.84 & 185 & 46.35 & 18.22 & 41.79 & 165 & 57.55 & 23.68 & 53.92 & 639 \\
    \midrule
    \multicolumn{14}{c}{\textbf{\textsc{\modelname} - content guidance}} \\
    \midrule
    Oracle & Lead & 48.19 & 19.99 & 44.25 & 205 & 49.61 & 19.27 & 44.75 & 164 & 60.47 & 24.93 & 56.89 & 649 \\
    Fixed & Lead & 46.10 & 19.21 & 42.43 & 202 & 47.98 & 18.99 & 43.41 & 165 & \underline{59.19} & \underline{24.72} & \underline{55.78} & 649 \\
    Fixed & Model & \underline{\textbf{47.20}} & \underline{\textbf{20.17}} & \underline{\textbf{43.48}} & 197 & \underline{\textbf{49.16}} & \underline{\textbf{20.17}} & \underline{\textbf{44.59}} & 164 & \underline{\textbf{60.00}} & \underline{\textbf{25.33}} & \underline{\textbf{56.49}} & 648 \\
    Model & Model & \underline{47.00} & \underline{20.15} & \underline{43.19} & 179 & 48.17 & \underline{20.06} & 44.07 & 164 & \underline{59.35} & \underline{25.25} & \underline{55.84} & 641 \\
    \bottomrule
  \end{tabular}
  \caption{ROUGE F1 scores and average words per summary on the \textbf{validation sets} for different types of guidance during inference. \emph{Lead} guidance is the first $k$ sentences from the source document (Section \ref{sec:results_content_guidance}). Model guidance is provided by BART-large for GovReport and BigBird for PubMed and arXiv. The choice of budget guidance values is described in Appendix \ref{sec:inference_details}. Results for models marked with $\dagger$ are taken from the original publications. \underline{Underlined results} are statistically  equivalent to the best methods ($p<0.05$).}\label{tab:experiment_results_validation}
\end{table*}

\section{Sample Summaries}
\label{sec:sample_summaries}
In Tables \ref{tab:summary_sample_pubmed1} to \ref{tab:summary_sample_govreport2}, we provide samples of summaries from PubMed, arXiv, and GovReport test sets. We compare BigBird and BART summaries to \textsc{\modelname} with lead and model-provided content guidance. Budget guidance is fixed to the same values as described in Section \ref{sec:results}.

\begin{table*}[ht]
  \centering
  \begin{tabular}{p{0.9\linewidth}}
    \toprule
    \multicolumn{1}{c}{\textbf{Reference abstract} (Tokens: 156)} \\
    \midrule
    we report a 2.5-year - old girl who presented with hoarseness of voice since 3 months of age and failure to thrive . chest x - ray showed cardiomegaly with a deviation of the trachea and mediastinum to the right side . two - dimensional echocardiography showed decreased flow across the right pulmonary artery , a small atrial septal defect ( asd ) with a right - to - left shunt , and a dilated right atrium and right ventricle with severe tricuspid regurgitation suggestive of severe pulmonary hypertension . a silent large patent ductus arteriosus was also seen . multiple detector computerized tomography aortogram confirmed the findings of absent right pulmonary artery and hypoplastic right lung with small cystic lesions suggestive of congenital cystic adenomatoid malformation in the right lower lobe .\\
    \midrule
    \multicolumn{1}{c}{\textbf{BigBird summary}} \\
    \multicolumn{1}{c}{Tokens: 48; ROUGE-1: 45.16; ROUGE-2: 31.52; ROUGE-L: 39.78} \\
    \midrule
    we report a 2.5-year - old girl with congenital absence of the right pulmonary artery with associated congenital cystic adenomatoid malformation of the right lower lobe, patent ductus arteriosus, and atrial septal defect, who presented with ortner's syndrome due to severe pulmonary hypertension. .\\
    \midrule
    \multicolumn{1}{c}{\textbf{\textsc{\modelname} - Fixed budget and Lead content guidance}} \\
    \multicolumn{1}{c}{Tokens: 199; ROUGE-1: 47.17; ROUGE-2: 26.58; ROUGE-L: 44.65} \\
    \midrule
     unilateral absence of pulmonary artery ( uapa ) is a rare congenital abnormality , with an estimated prevalence of 1 in 200,000. while some patients with uapa are totally asymptomatic , others may have severe pulmonary hypertension. we report a 2.5-year - old girl with congenital absence of the right pulmonary artery with associated congenital cystic adenomatoid malformation ( ccam ) , patent ductus arteriosus ( pda ) , and atrial septal defect , who presented with ortner 's syndrome due to severe pulmonary hypertension. a 2.5-year - old girl presented with hoarseness of voice noticed since 3 months of age , breathlessness for the past 15 days and failure to thrive. pulmonary agenesis is usually unilateral , right sided absence of pulmonary artery being more common. the syndrome has since been described in adults with various cardiovascular disorders , but reports in children are less common. congenital cystic adenomatoid malformation ( ccam ) is a rare cause of congenital cyanotic heart disease. the child was intubated and ventilated and started on pressors , but sustained a cardiac arrest on the 4th hospital day from which she could not be resuscitated.\\
    \midrule
    \multicolumn{1}{c}{\textbf{\textsc{\modelname} - Fixed budget and BigBird content guidance}} \\
    \multicolumn{1}{c}{Tokens: 142; ROUGE-1: 42.42; ROUGE-2: 23.66; ROUGE-L: 38.64} \\
    \midrule
    congenital cystic adenomatoid malformation ( ccam ) is a rare cause of congenital cyanotic heart disease. pulmonary agenesis is usually unilateral , right sided absence of pulmonary artery being more common. the syndrome has since been described in adults with various cardiovascular disorders , but reports in children are less common. unilateral absence of pulmonary artery ( uapa ) is a rare congenital abnormality , with an estimated prevalence of 1 in 200,000. we report a 2.5-year - old girl with congenital absence of the right pulmonary artery with associated congenital cystic adenomatoid malformation ( ccam ) , patent ductus arteriosus ( pda ) , and atrial septal defect, who presented with ortner 's syndrome due to severe pulmonary hypertension. while some patients with uapa are totally asymptomatic , others may have severe pulmonary hypertension.\\
    \bottomrule
  \end{tabular}
  \caption{Sample abstract and generated summaries from the PubMed test set (ID = 5836).} \label{tab:summary_sample_pubmed1}
\end{table*}

\begin{table*}[ht]
  \centering
  \begin{tabular}{p{0.95\linewidth}}
    \toprule
    \multicolumn{1}{c}{\textbf{Reference abstract} (Tokens: 240)} \\
    \midrule
    the navier - stokes - fourier theory of viscous , heat - conducting fluids provides parabolic equations and thus predicts infinite pulse speeds . naturally this feature has disqualified the theory for relativistic thermodynamics which must insist on finite speeds and , moreover , on speeds smaller than c. the attempts at a remedy have proved heuristically important for a new systematic type of thermodynamics : extended thermodynamics . that new theory has symmetric hyperbolic field equations and thus it provides finite pulse speeds.extended thermodynamics is a whole hierarchy of theories with an increasing number of fields when gradients and rates of thermodynamic processes become steeper and faster . the first stage in this hierarchy is the 14-field theory which may already be a useful tool for the relativist in many applications . the 14 fields  and further fields  are conveniently chosen from the moments of the kinetic theory of gases.the hierarchy is complete only when the number of fields tends to infinity . in that case. the pulse speed of non - relativistic extended thermodynamics tends to infinity while the pulse speed of relativistic extended thermodynamics tends to c , the speed of light. [...]\\
    \midrule
    \multicolumn{1}{c}{\textbf{BigBird summary}} \\
    \multicolumn{1}{c}{Tokens: 105; ROUGE-1: 37.27; ROUGE-2: 13.75; ROUGE-L: 32.29} \\
    \midrule
    the paradox of pulse speeds in extended thermodynamics has been known for 50 years. it seems to have been caused by eckart s theory of irreversible processes which assumed a constant pulse speed and a fixed temperature. recently it has been shown by boillat \& ruggeri that, as the number of moments increases, the pulse speed tends to infinity in the non - relativistic kinetic theory of gases and the relativistic case by which the pulse speed tends to c. these results put an end to the long - standing paradox of pulse speeds. they are reviewed in detail.\\
    \midrule
    \multicolumn{1}{c}{\textbf{\textsc{\modelname} - fixed budget and Lead content guidance}} \\
    \multicolumn{1}{c}{Tokens: 216; ROUGE-1: 50.12; ROUGE-2: 14.18; ROUGE-L: 44.77} \\
    \midrule
    the pulse speed problem is one of the most important questions in thermodynamics , but it is a question that can be answered , and has to be answered. we derive a set of thermodynamic processes for a , a , b , c , and d. in this paper , we prove that a co - vector exists in which the entropy is a function of the thermal equation of the state. the thermodynamics of viscous , heat - conducting gases is studied by means of the determination of the 14 fields   of the field equations. in this paper , we review the recent developments in the field of non - equilibrium thermodynamics. it is possible , and indeed common , to make a specific choice for the fields u and the concavity postulate is contingent upon that choice. the first moments in the kinetic theory of gases are obtained from a homogeneous system   where the acceleration waves and their speeds of propagation are to be calculated from the homogeneous systems. the heat fluxes f(x , p , t ) of the atoms , viz. the paper deals with the thermodynamics of a non - degenerate gas. in the non - relativistic limit , \\
    \midrule
    \multicolumn{1}{c}{\textbf{\textsc{\modelname} - fixed budget and BigBird content guidance}} \\
    \multicolumn{1}{c}{Tokens: 190; ROUGE-1: 44.67; ROUGE-2: 12.75; ROUGE-L: 42.64} \\
    \midrule
    it is then a simple problem of linear algebra to prove that the entropy density h = ha is concave as a function of f. the pulse speed problem is one of the most important questions in thermodynamics , but it is a question that can be answered , and has to be answered. it is possible , and indeed common , to make a specific choice for the fields u and the concavity postulate is contingent upon that choice. the first moments in the kinetic theory of gases are obtained from a homogeneous system   where the acceleration waves and their speeds of propagation are to be calculated from the homogeneous systems. we derive a set of thermodynamic processes for a , a , b , c , and d. in this paper , we review the recent developments in non - relativistic kinetic theory. this paper presents the results of an investigation of the non - degenerate gases in the system. we conclude that no paradox of infinite speeds can arise in extended thermodynamics. in the non - relativistic limit , \\
    \bottomrule
  \end{tabular}
  \caption{Sample abstract and generated summaries from the PubMed test set (ID = 1475).} \label{tab:summary_sample_pubmed2}
\end{table*}

\begin{table*}[ht]
  \centering
  \begin{tabular}{p{0.9\linewidth}}
    \toprule
    \multicolumn{1}{c}{\textbf{Reference abstract} (Tokens: 157)} \\
    \midrule
    the metrization of the space of neural responses is an ongoing research program seeking to find natural ways to describe , in geometrical terms , the sets of possible activities in the brain . one component of this program are the \_ spike metric , notions of distance between two spike trains recorded from a neuron . alignment spike metrics work by identifying `` equivalent '' spikes in one train and the other . we present an alignment spike metric having @xmath0 underlying geometrical structure ; the @xmath1 version is euclidean and is suitable for further embedding in euclidean spaces by multidimensional scaling methods or related procedures . we show how to implement a fast algorithm for the computation of this metric based on bipartite graph matching theory . @xmath2center for studies in physics and biology , rockefeller university , new york ny @xmath3harvard faculty of arts and sciences , cambridge ma\\
    \midrule
    \multicolumn{1}{c}{\textbf{BigBird summary}} \\
    \multicolumn{1}{c}{Tokens: 89; ROUGE-1: 28.71; ROUGE-2: 3.00; ROUGE-L: 22.77} \\
    \midrule
    we propose a spike metric that is consistent with the time - coding hypothesis of spike generation and has all of the desirable properties of an @xmath0 norm . when @xmath1 , this metric is equal to the victor - purpura metric @xcite . when @xmath2 , this metric is equal to the hungarian algorithm @xcite , @xcite , @xcite . when @xmath3 , this metric is equal to the monge algorithm @xcite , @xcite , @xcite .\\
    \midrule
    \multicolumn{1}{c}{\textbf{\textsc{\modelname} - Fixed budget and Lead content guidance}} \\
    \multicolumn{1}{c}{Tokens: 165; ROUGE-1: 42.51; ROUGE-2: 6.32; ROUGE-L: 38.33s} \\
    \midrule
    we consider the problem of finding a minimum weight matching on a bipartite graph such that each edge of the graph has a weight or cost. the value of the metric between two spike trains is the @xmath0 norm of the difference between their estimated rate functions . this metric preserves the integrity of individual spikes instead of viewing them as contributions to a rate function . we present a spike metric satisfying two important desiderata : that it be grounded in the time - coding hypothesis of spike generation , and that it is closely related to the euclidean @xmath1 norm . our proof uses the tools of graph theory and combinatorial optimization , and demonstrates that they can be usefully brought to bear on problems in computational biology. the cost of aligning two spikes on different trains is @xmath0. we present a new metric for comparing two spike trains , which we call the hungarian metric.\\
    \midrule
    \multicolumn{1}{c}{\textbf{\textsc{\modelname} - Fixed budget and BigBird content guidance}} \\
    \multicolumn{1}{c}{Tokens: 150; ROUGE-1: 41.61; ROUGE-2: 5.88; ROUGE-L: 38.69} \\
    \midrule
    we present a spike metric satisfying two important desiderata : that it be grounded in the time - coding hypothesis of spike generation , and that it is closely related to the euclidean @xmath1 norm . the value of the metric between two spike trains is the @xmath0 norm of the difference between their estimated rate functions . the hungarian algorithm is a special case of algorithms to solve general matching problems and assignment problems. the metric is equal to the sum of the costs incurred by the most efficient alignment. we consider the problem of finding a minimum weight matching on a bipartite graph such that each edge of the graph has a weight or cost. we present a new metric for comparing two spike trains , which we call the hungarian metric. our metric is a minimization over all possible matchings .\\
    \bottomrule
  \end{tabular}
  \caption{Sample abstract and generated summaries from the arXiv test set (ID = 5946).} \label{tab:summary_sample_arxiv1}
\end{table*}

\begin{table*}[ht]
  \centering
  \begin{tabular}{p{0.95\linewidth}}
    \toprule
    \multicolumn{1}{c}{\textbf{Reference abstract} (Tokens: 145)} \\
    \midrule
    in this paper we discuss the chemical evolution of elliptical galaxies and its consequences on the evolution of the intracluster medium ( icm ) . we use chemical evolution models taking into account dark matter halos and compare the results with previous models where dark matter was not considered . in particular , we examine the evolution of the abundances of some relevant heavy elements such as oxygen , magnesium and iron and conclude that models including dark matter halos and an initial mass function ( imf ) containing more massive stars than the salpeter ( 1955 ) imf , better reproduce the observed abundances of mg and fe both in the stellar populations and in the icm ( asca results ) . we also discuss the origin of gas in galaxy clusters and conclude that most of it should have a primordial origin .\\
    \midrule
    \multicolumn{1}{c}{\textbf{BigBird summary}} \\
    \multicolumn{1}{c}{Tokens: 109; ROUGE-1: 36.21; ROUGE-2: 9.57; ROUGE-L: 28.45} \\
    \midrule
    we discuss a model for the chemical evolution of elliptical galaxies in which supernovae ( sne)-driven galactic winds play an important role in the formation of these objects . in this model  the star formation is assumed to stop after the occurrence of the galactic wind and the galaxy evolves passively thereafter . the star formation is assumed to stop after the occurrence of the galactic wind and the galaxy evolves passively thereafter . the model includes the most recent ideas on sn progenitors and nucleosynthesis , indicating that sne ia originate from long living stars whereas sne of type ii originate from short living stars .\\
    \midrule
    \multicolumn{1}{c}{\textbf{\textsc{\modelname} - fixed budget and Lead content guidance}} \\
    \multicolumn{1}{c}{Tokens: 159; ROUGE-1: 43.73; ROUGE-2: 12.27; ROUGE-L: 39.43} \\
    \midrule
    this is a very strong conclusion since it implies a very fast process for the formation of big ellipticals at variance with the hierarchical clustering scenario for galaxy formation. we show how abundance ratios in stellar populations and gas in ellipticals can be used to constrain the amount and concentration of dark matter in these objects. in order to reproduce realistic galaxies , namely with the right colors and luminosities. we discuss the chemical evolution of elliptical galaxies in the framework of a simple model based on the idea that the efficiency of star formation should be inversely proportional to the dynamical timescale. we find that the efficiency of star formation in elliptical galaxies is inversely proportional to the dynamical timescale. in particular , we show that it is not possible to explain the increase of the [ mg / fe ] ratio in the nuclei of ellipticals as a function of galactic luminosity.\\
    \midrule
    \multicolumn{1}{c}{\textbf{\textsc{\modelname} - fixed budget and BigBird content guidance}} \\
    \multicolumn{1}{c}{Tokens: 147; ROUGE-1: 43.70; ROUGE-2: 14.93; ROUGE-L: 39.26} \\
    \midrule
    we discuss the chemical evolution of elliptical galaxies in the framework of a simple model based on the idea that the efficiency of star formation should be inversely proportional to the dynamical timescale. in particular , we show that the efficiency of star formation increases with the total mass of the galaxy and that the more massive galaxies develop a galactic wind before the less massive ones. we show that the presence of dark matter in elliptical galaxies plays a crucial role in determining the onset and the entity of galactic winds. in this paper we discuss the possibility of an inverse wind scenario for the formation of elliptical galaxies. the model includes the most recent ideas on sn progenitors and nucleosynthesis , indicating that sne ia originate from long living stars whereas sne of type ii originate from short living stars.\\
    \bottomrule
  \end{tabular}
  \caption{Sample abstract and generated summaries from the arXiv test set (ID = 6213).} \label{tab:summary_sample_arxiv2}
\end{table*}

\begin{table*}[ht]
  \centering
  \begin{tabular}{p{0.95\linewidth}}
    \toprule
    \multicolumn{1}{c}{\textbf{Reference abstract} (Tokens: 474)} \\
    \midrule
    Congress frequently faces questions about whether and how to commemorate people and events that have influenced the nation's history. Congress often has chosen to do so by establishing national memorials or by conferring a national designation on existing state, local, or private memorials. The National Park Service (NPS) defines national memorials within the National Park System as "primarily commemorative" works that need not be at sites historically associated with their subjects. The Commemorative Works Act (CWA; 40 U.S.C. §§8901-8910) was enacted to govern the establishment process for memorials located in the District of Columbia (Washington, DC) or its environs that are under the jurisdiction of the NPS or the General Services Administration. The CWA includes provisions related to memorial location, design, construction, and perpetual maintenance. Memorials in Washington, DC, include those with the word national in the name and those that are essentially national memorials but do not bear that title. For memorials outside the District of Columbia, no specific law or set of regulations governs their establishment. Congress has established a number of federally administered national memorials throughout the nation, most often as units of the National Park System but also under management of other federal agencies. Various nonfederal entities undertaking commemorative efforts also have petitioned Congress for assistance or statutory recognition, and some individual memorial organizers have titled their works as national memorials without congressional recognition. To clarify options for Congress when considering commemoration of individuals, groups, and events through memorials, this report discusses several types of congressional involvement in memorials outside the District of Columbia. For purposes of the report, these are characterized as high federal involvement (e.g., congressional establishment of a national memorial under federal agency administration); medium federal involvement (e.g., congressional authorization for a memorial to be located on federal property or to receive federal funds); low federal involvement (e.g., statutory recognition without additional federal support); and no federal involvement (e.g., a self-declared national memorial). The report provides examples of memorials of each type and discusses some options for Congress, with regard to both individual memorial designations and consideration of whether to systematize criteria for memorials outside Washington, DC, similar to the CWA's provisions for District of Columbia memorials. Because this report focuses specifically on memorials outside the District of Columbia, please see CRS Report R41658, Commemorative Works in the District of Columbia: Background and Practice, by Jacob R. Straus, for discussion of memorials governed by the CWA in Washington, DC, and its environs.\\
    \bottomrule
  \end{tabular}
  \caption{Sample abstract from the GovReport test set (ID = 681). Model predictions are presentend in the next pages.} \label{tab:abstract_sample_govreport}
\end{table*}

\begin{table*}[ht]
  \centering
  \begin{tabular}{p{0.95\linewidth}}
    \toprule
    \multicolumn{1}{c}{\textbf{BART-large summary}} \\
    \multicolumn{1}{c}{Tokens: 390; ROUGE-1: 56.11; ROUGE-2: 19.38; ROUGE-L: 54.23} \\
    \midrule
    Virtually all societies attempt to remember and memorialize individuals, groups, and events as part of the preservation of shared rhetoric and history. In the United States, there are hundreds, and possibly thousands, of commemorative works to commemorate such events. Decisions about which places to commemorate are made by many different entities, including Congress, federal agencies, state and local governments, and private citizens, among others. Some memorials receive federal involvement, while others do not. For example, for memorials on federal land in the District of Columbia, the Commemorative Works Act (CWA) requires that Congress authorize the creation of a new memorial. No systematic law or set of regulations governs the establishment of memorials outside Washington, DC. However, Congress also has established or recognized numerous memorials nationwide, and some have been designated by the executive branch. Federal agencies may be classified as "high," "medium," "low," or "none." Memorials with high federal involvement typically are located on U.S. federal land; receive federal funds for design, construction, and maintenance; and are managed by federal agencies. These include memorials established by Congress as units of the National Park System or under the administration of another agency. Other memorials with no federal involvement are those that do not receive any direct federal involvement (i.e., memorials designated by Congress but not administered by a federal agency). Congress, executive branch officials, and other interested parties may place plaques, memorials, and similar works at federal sites in remembrance of a person, group, or event. or on nonfederal land. The National Park Service (NPS) and the General Services Administration (GSA) maintain some of the nation's largest memorials. NPS provides assistance to other federal agencies with assistance in managing memorials located on its lands, including NPS-designated "NPS-affiliated areas." Other agencies, primarily the Army and Air Force, have similar relationships with NPS affiliated areas. Congress has appropriated funds to both NPS and GSA to help fund memorials created on the lands they manage.\\
    \bottomrule
  \end{tabular}
  \caption{Summary generated by BART for a document from GovReport test set (ID = 681). Reference summary is presented in Table \ref{tab:abstract_sample_govreport}.} \label{tab:summary_sample_govreport}
\end{table*}

\begin{table*}[ht]
  \centering
  \begin{tabular}{p{0.95\linewidth}}
    \toprule
    \multicolumn{1}{c}{\textbf{\textsc{\modelname} - fixed budget and Lead content guidance}} \\
    \multicolumn{1}{c}{Tokens: 548; ROUGE-1: 63.37; ROUGE-2: 24.77; ROUGE-L: 57.30} \\
    \midrule
    Beyond these federally endorsed memorials, a wide variety of other entities have established and maintained memorials throughout the country with no federal connection, including some titled as "national memorials,"In the United States, there are hundreds, and possibly thousands, of memorials to various individuals, groups, and events. Decisions about which people, groups, or events to memorialize are made by many different entities, including Congress, federal agencies, state and local governments, and private citizens, among others. For example, the CWA governs the establishment of memorials on federal lands in the District of Columbia, with provisions for the creation, design, construction, and maintenance of such works. In other areas, various laws, regulations, and policies may provide for different groups and governments to decide what should be commemorated and how. For certain types of commemorations, Congress has taken a more systematized approach. No systematic law or set of regulations governs the establishment of memorials outside Washington, DC. ,Some of these memorials include multiple facilities such as a visitor center or kiosk in addition to the primary commemor For example, the George Washington Masonic National Memorial in Alexandria, VA, and the National Memorial for Peace and Justice in Montgomery, AL, are privately established and maintained. In some cases, memorials located outside of the District of Columbia have been called "national" memorials without being so designated by Congress, such as through the establishment of a program to identify nonfederal memorials deserving of a national designation. A distinction is drawn between memorials located within and outside of Washington, DC, because of the exclusive role the CWA gives Congress to authorize new memorials on federal land in the District of Columbia, and the role of federal agencies—primarily the National Park Service (NPS) and the General Services Administration (GSA)—in maintaining District-based memorials once dedicated. This report considers the extent of federal involvement in memorials located outside the District of Columbia. Congress also could potentially consider a program to provide grants to nonfederal entities for constructing and/or maintaining national memorials outside of Washington, DC. While many such works are established without federal involvement, Congress also has established or recognized numerous memorials nationwide, and some have been designated by the executive branch. For purposes of this report, federal involvement in memorials outside the District of Columbia may be classified as "high," "medium," "low," or "none." For example, P. L. Other variations of federal-nonf For a discussion of the process for creating a new NPS unit and associated issues, see CRS Report RS20158, National Park System: Establishing New Units. Legislation designating these national memorials often includes explicit language stating that the memorial is not an NPS unit and that federal funds shall not be provided for the memorial. In some instances, Congress authorizes a memorial to be created on federal land and administered by a federal agency.\\
    \bottomrule
  \end{tabular}
  \caption{Summary generated by \textsc{\modelname} with BART content guidance for a document from GovReport test set (ID = 681). Reference summary is presented in Table \ref{tab:abstract_sample_govreport}.} \label{tab:summary_sample_govreport_lead_guidance}
\end{table*}

\begin{table*}[ht]
  \centering
  \begin{tabular}{p{0.95\linewidth}}
    \toprule
    \multicolumn{1}{c}{\textbf{\textsc{\modelname} - fixed budget and BART-large content guidance}} \\
    \multicolumn{1}{c}{Tokens: 548; ROUGE-1: 63.37; ROUGE-2: 24.77; ROUGE-L: 57.30} \\
    \midrule
    In other areas, various laws, regulations, and policies may provide for different groups and governments to decide what should be commemorated and how. Beyond these federally endorsed memorials, a wide variety of other entities have established and maintained memorials throughout the country with no federal connection, including some titled as "national memorials,"In the United States, there are hundreds, and possibly thousands, of memorials to various individuals, groups, and events. Decisions about which people, groups, or events to memorialize are made by many different entities, including Congress, federal agencies, state and local governments, and private citizens, among others. This report considers the extent of federal involvement in memorials located outside the District of Columbia. For example, the CWA governs the establishment of memorials on federal lands in the District of Columbia, with provisions for the creation, design, construction, and maintenance of such works. No systematic law or set of regulations governs the establishment of memorials outside Washington, DC. In some cases, memorials located outside of the District of Columbia have been called "national" memorials without being so designated by Congress, such as through the establishment of a program to identify nonfederal memorials deserving of a national designation. For purposes of this report, federal involvement in memorials outside the District of Columbia may be classified as "high," "medium," "low," or "none." Legislation designating these national memorials often includes explicit language stating that the memorial is not an NPS unit and that federal funds shall not be provided for the memorial. For a discussion of the process for creating a new NPS unit and associated issues, see CRS Report RS20158, National Park System: Establishing New Units. While many such works are established without federal involvement, Congress also has established or recognized numerous memorials nationwide, and some have been designated by the executive branch. For certain types of commemorations, Congress has taken a more systematized approach. Congress also could potentially consider a program to provide grants to nonfederal entities for constructing and/or maintaining national memorials outside of Washington, DC. A distinction is drawn between memorials located within and outside of Washington, DC, because of the exclusive role the CWA gives Congress to authorize new memorials on federal land in the District of Columbia, and the role of federal agencies—primarily the National Park Service (NPS) and the General Services Administration (GSA)—in maintaining District-based memorials once dedicated. ,Some of these memorials include multiple facilities such as a visitor center or kiosk in addition to the primary commemor For example, the George Washington Masonic National Memorial in Alexandria, VA, and the National Memorial for Peace and Justice in Montgomery, AL, are privately established and maintained. In some instances, Congress authorizes a memorial to be created on federal land and administered by a federal agency. For example, P. L. Other variations of federal-nonf\\
    \bottomrule
  \end{tabular}
  \caption{Summary generated by \textsc{\modelname} with BART content guidance for a document from GovReport test set (ID = 681). Reference summary is presented in Table \ref{tab:abstract_sample_govreport}. Note that this summary uses the same set of summary views as \textsc{\modelname} with Lead content guidance in Table \ref{tab:summary_sample_govreport_lead_guidance}, just changing their presentation order.} \label{tab:summary_sample_govreport2}
\end{table*}

\end{document}